\documentclass[preprint,12pt,authoryear]{elsarticle}
\usepackage{amsmath}
\usepackage{enumitem}
\usepackage{graphicx}
\usepackage{floatrow}
\usepackage{placeins}
\usepackage{booktabs}
\usepackage{subcaption}
\usepackage[ruled,longend,linesnumbered] {algorithm2e}
\usepackage{xcolor}
\usepackage{color}
\usepackage{tabularx}
\usepackage[font=small,labelfont=bf,labelsep=period]{caption}
\usepackage{amssymb}
\usepackage{booktabs}
\usepackage{tabularx}
\usepackage{makecell}
\usepackage[colorlinks=true, linkcolor=blue, citecolor=blue]{hyperref}
\usepackage{multirow}  
\hypersetup{pdftitle={A Differentiable Composite Approximation Framework for Autonomous Underwater Vehicle Maneuvering Modeling from Sea-Trial Data},
            pdfauthor={Aobo Wang, Aifei Xia, Zihao Wang, Lizhu Hao, Shaorong Xie}}
\myfooterfont{\scriptsize}
\myfooter[L]{Accepted manuscript, \textit{Ocean Engineering}. \href{https://doi.org/10.1016/j.oceaneng.2026.129398}{doi:10.1016/j.oceaneng.2026.129398}. \href{https://creativecommons.org/licenses/by-nc-nd/4.0/}{CC BY-NC-ND 4.0}.}

\begin{document}
\begin{frontmatter}

\title{A Differentiable Composite Approximation Framework for Autonomous Underwater Vehicle Maneuvering Modeling from Sea-Trial Data}
\author[label1,label2]{Aobo Wang}
\author[label3]{Aifei Xia}
\author[label3]{Zihao Wang\corref{cor1}}
\author[label4]{Lizhu Hao}
\author[label5]{Shaorong Xie}
\affiliation[label1]{
			organization={College of Shipbuilding Engineering, Harbin Engineering University},
            city={Harbin},
            postcode={150001}, 
            country={China}}
\affiliation[label2]{
			organization={China Academy of Aerospace Aerodynamics},
			city={Beijing},
            postcode={100074}, 
            country={China}}
\affiliation[label3]{
			organization={School of Future Technology, Shanghai University},
			city={Shanghai},
            postcode={200444}, 
            country={China}}   
\affiliation[label4]{
			organization={China Ship Scientific Research Center},
			city={Wuxi},
            postcode={214082}, 
            country={China}}
\affiliation[label5]{
			organization={School of Mechanical Engineering and Automation, Shanghai University},
			city={Shanghai},
            postcode={200444}, 
            country={China}}  
\cortext[cor1]{Corresponding author: zihaowang@shu.edu.cn}

\begin{abstract}
\small
Field-based modeling from onboard measurements can produce autonomous underwater vehicle (AUV) maneuvering models that reflect real operating characteristics. From an approximation perspective, conventional maneuvering models use predefined constraint polynomial bases, whereas data-driven models use data-adaptive bases. Motivated by this basis-function view, this paper presents a differentiable composite approximation formulation, in which the polynomial basis component and the data-adaptive basis component are treated as differentiable parts of a unified predictor and calibrated jointly. A gradient-based co-calibration method is developed for full-scale AUV maneuvering prediction, where a sensitivity-aware mechanism regulates bounded polynomial updates while the neural residual captures remaining nonlinear discrepancies under a shared prediction objective. To account for ocean-current effects in field data, a turning-motion-based current estimation and compensation procedure is incorporated to construct current-compensated learning targets for training and rollout. The framework is evaluated using sea-trial data collected from a 7-meter AUV under multiple maneuvering conditions. Results show that the proposed method improves recursive trajectory and velocity prediction compared with polynomial-only, neural-only, and frozen-prior hybrid baselines, demonstrating its applicability to field-data-based AUV maneuvering modeling.
\end{abstract}
\end{frontmatter}

\section{Introduction}
Autonomous Underwater Vehicles (AUVs) play a vital role in marine applications such as exploration, monitoring, and defense. As missions grow in complexity, AUVs are expected to operate autonomously over extended periods with minimal human intervention. Recent studies have explored learning and adaptive control for autonomous marine applications \citep{wang2025acousticservo,li2022dynamicpositioning} and nonlinear systems subject to practical constraints \citep{naderolasli2026pendulum,naderolasli2026piezo,naderolasli2026gimbal}. To support such capabilities, precise dynamic modeling becomes essential. By capturing the nonlinear characteristics of underwater motion, these models form the foundation for the agent to understand how it interacts with its environment and serve as a key enabler for reliable state prediction, informed decision-making, model-aided navigation, and model-based control.

Conventional AUV dynamic models are often derived from polynomial approximations of hydrodynamic forces and moments, where predefined polynomial basis functions involving velocity states and control inputs are fitted to experimental data. The unknown parameters of the polynomial basis terms, called hydrodynamic coefficients, are relatively interpretable and compact, making them suitable for control design. Common approaches for estimating these coefficients include captive model tests \citep{nouri2016}, CFD simulations \citep{du2018, go2019}, and free-running based system identification methods \citep{ahmed2023}. 

Among these methods, free-running based system identification methods are more suitable for in-field modeling and post-deployment recalibration, as they do not require specialized facilities or computational resources. They estimate the coefficients by fitting the model output to measured free-running data. Prior studies have identified surge or yaw dynamics using least squares \citep{hegrenaes2007, zhang2020}, Kalman filtering \citep{tiano2007, sabet2017, busse2022}, and H-infinity filter \citep{sajedi2019}. According to established maneuvering theory, third-order polynomial basis functions are commonly adopted to represent the dominant nonlinear hydrodynamic effects of marine vehicles \citep{fossen2021}. Such polynomial models provide sufficient approximation when the selected terms are well conditioned and sufficiently excited by the available data. However, as the number of hydrodynamic derivatives increases, especially in coupled multi-degree-of-freedom motion, the regression basis may become highly correlated. This collinearity can lead to parameter drift and poor identifiability: the model may fit the measured trajectories in a mathematical sense, while the estimated coefficients become physically less reliable and the resulting predictor generalizes poorly outside the calibration data.

Learning-based data-driven modeling methods provide another route for approximating vehicle dynamics, which can be interpreted as data-adaptive basis function approximation from the perspective of approximation theory. Instead of prescribing a fixed set of polynomial basis functions, these methods learn implicit basis representations from data. For example, neural networks construct highly flexible function approximators through cascaded nonlinear transformations, enabling them to capture complex nonlinear relationships and interaction effects without explicitly specifying the functional form in advance. Related approaches have been explored for marine vehicle dynamics, including multi-output Gaussian processes \citep{ramirez2021}, ensemble MLP models \citep{Tesnar2025}, and physics-informed neural networks \citep{zhao2024}. \citet{ramirez2021} employed multi-output Gaussian processes to model the dynamics of an underactuated AUV with limited data, while \citet{Tesnar2025} constructed an ensemble-MLP dynamics model and used predictive uncertainty to guide data selection for rehearsal in storage-limited online learning. Physics-informed neural networks have also been introduced to incorporate governing equations into neural prediction models \citep{zhao2024,an2025pinnexcitation}.

However, the strong representation capacity of these adaptive approximators is a double-edged sword. Although they offer high flexibility for modeling complex and high-dimensional dynamics, they may also overfit spurious correlations in the training data, especially when the available sea-trial data are limited and noisy. Such correlations may not correspond to physically meaningful dynamics and may become unreliable when the vehicle operates outside the training distribution.

These limitations motivate physics-guided or hybrid physics-data modeling strategies. In a typical pipeline, a polynomial basis maneuvering model provides a structured hydrodynamic backbone, while a neural component compensates for the effects that are not captured by the polynomial model. This idea is well known as grey-box modeling in the field of system identification. For example, \cite{lei2022pgnn} used a physics-guided neural network in which a theoretical model provides the baseline and an RBF neural network compensates for modeling errors. Incremental residual learning has also been investigated for underwater vehicles to compensate for the physical-modeling gap caused by environmental uncertainty \citep{lei2022incremental}. Similar hybrid ideas have also been explored in marine ship-motion prediction and digital-twin modeling \citep{skulstad2021, nielsen2022}. Nevertheless, these grey-box modeling methods commonly follow a sequential workflow: the mechanistic component is first assumed to provide a sufficiently reliable baseline and is then fixed, while a neural network is trained to compensate for the remaining error. 
If an imperfect polynomial model is first identified from free-running data and then frozen, its coefficients may absorb effects outside the selected model structure, including environmental disturbances, higher-order nonlinearities, and omitted hydrodynamic couplings. Such a biased baseline may fit the training data but yield coefficients with reduced physical reliability and interpretability, forcing the neural residual to compensate for both unmodeled dynamics and baseline error, which may degrade recursive maneuvering prediction. 

From the perspective of function approximation, polynomial and neural models represent two distinct approaches to approximating unknown dynamical mappings. Polynomial models construct the approximation using prescribed basis functions, whereas neural networks learn adaptive representations from data. Their complementary approximation characteristics suggest that combining these representations may better balance model structure and flexibility in AUV maneuvering dynamics. Building on this interpretation, this study moves beyond viewing hybrid modeling solely as a combination of mechanistic and neural components, and instead formulates it as a composite approximation problem involving complementary function approximation representations. This unified perspective can provide a flexible foundation for organizing and coupling different approximation components within a unified predictive structure.

Based on this approximation-theoretic perspective, this study develops a differentiable composite approximation framework for AUV maneuvering modeling using full-scale sea-trial data. In this formulation, the polynomial and data-adaptive basis components are integrated as differentiable parts of a single transition predictor and optimized under a common prediction loss, enabling joint calibration rather than separate-stage fitting. A sensitivity-aware gradient-based calibration mechanism is developed in which the polynomial coefficients remain adjustable within bounded ranges instead of being fixed as baseline parameters, while the data-adaptive basis captures the remaining nonlinear discrepancies. To reduce the influence of ocean currents on model learning, a turning-motion-based current estimation and compensation procedure is further incorporated. Validation using sea-trial data collected from a 7-meter AUV shows that the proposed co-calibration framework enables simultaneous optimization of both basis components and improves field-data-based AUV maneuvering prediction.

The remainder of this paper is organized as follows: Section 2 formulates the maneuvering modeling problem and the joint calibration objective. Section 3 presents the proposed modeling method. Section 4 provides experimental validation and analysis, while Section 5 concludes the paper.

\section{Problem Formulation}
The objective of this study is to construct an AUV maneuvering dynamics model from in-field free-running sea-trial data. In sea trials, the measured velocity is affected by ambient ocean currents. To represent hydrodynamic forces in terms of the vehicle velocity relative to the surrounding water, the measured velocity is decomposed into water-relative and current-induced components before model calibration. The learning target is therefore defined in the water-relative velocity frame, while the current component is reintroduced during rollout for comparison with sea-trial measurements. Under this formulation, the modeling problem is to jointly calibrate a polynomial hydrodynamic component and a neural residual component within a single predictive model.

\subsection{Horizontal-plane maneuvering data}

The AUV motion considered in this paper is restricted to horizontal-plane maneuvering. The inertial-frame pose and body-fixed velocity are defined as
\begin{equation}
\boldsymbol{\eta}_k =
\begin{bmatrix}
x_k & y_k & \psi_k
\end{bmatrix}^{\top},
\qquad
\boldsymbol{\nu}_k =
\begin{bmatrix}
u_k & v_k & r_k
\end{bmatrix}^{\top},
\label{eq:state_definition}
\end{equation}
where $x_k$ and $y_k$ are horizontal positions, $\psi_k$ is the heading angle, and $u_k$, $v_k$, and $r_k$ denote surge velocity, sway velocity, and yaw rate, respectively. The measured command input is written as
\begin{equation}
\boldsymbol{a}_k =
\begin{bmatrix}
n_k & \delta_k
\end{bmatrix}^{\top},
\label{eq:input_definition}
\end{equation}
where $n_k$ is the propeller revolution rate and $\delta_k$ is the equivalent rudder angle.

Let $\{\mathcal{D}_i\}_{i=1}^{N}$ be a set of free-running or quasi-open-loop maneuvering trials collected from the same vehicle. Each trial contains a time series
\begin{equation}
\mathcal{D}_i =
\left\{
\left(
\boldsymbol{\eta}^{(i)}_k,
\boldsymbol{\nu}^{(i)}_{m,k},
\boldsymbol{a}^{(i)}_k
\right)
\right\}_{k=1}^{T_i},
\label{eq:sea_trial_dataset}
\end{equation}
where $\boldsymbol{\nu}_{m,k}$ denotes the measured velocity available from onboard navigation data. 

The engineering objective is to obtain a maneuvering model that can be executed over a complete input sequence from the sea-trial data. In rollout prediction, the predicted velocity at one step is fed back into the model input for subsequent prediction.

\subsection{Velocity decomposition under current influence}

For maneuvering dynamics, the hydrodynamic response is more naturally associated with the vehicle velocity relative to the surrounding water than with the current-affected measured velocity. During a short sea-trial maneuver, the ambient current is approximated as locally constant in the inertial frame, with speed $V_c$ and direction $\beta_c$. Its projection into the body-fixed frame is
\begin{equation}
\boldsymbol{\nu}^{b}_{c,k}
=
\begin{bmatrix}
V_c\cos(\beta_c-\psi_k)\\
V_c\sin(\beta_c-\psi_k)\\
0
\end{bmatrix}.
\label{eq:current_projection_problem}
\end{equation}
The measured and water-relative velocities, both expressed in the body-fixed frame, are then related by
\begin{equation}
\boldsymbol{\nu}^{b}_{m,k}=\boldsymbol{\nu}^{b}_{r,k}+\boldsymbol{\nu}^{b}_{c,k},
\label{eq:relative_velocity_problem}
\end{equation}
where $\boldsymbol{\nu}^{b}_{r,k}=[u_{r,k},v_{r,k},r_k]^{\top}$ is the water-relative maneuvering velocity expressed in the body-fixed frame. For notational simplicity, the superscript $b$ is omitted in the following learning formulation, since measured and water-relative maneuvering velocities are expressed in the body-fixed frame unless otherwise stated. This transformation prevents the learned maneuvering model from mixing vehicle dynamics relative to the surrounding water with current-induced kinematic drift. The estimation of $V_c$ and $\beta_c$ from turning maneuvers is described in Section~3.

\subsection{\texorpdfstring{Modeling objective}{Modeling objective}}

After current compensation, the learning target is the one-step transition of the water-relative maneuvering velocity. Each sample is represented by the relative state-input vector
\begin{equation}
\boldsymbol{s}_{r,k}
=
\begin{bmatrix}
\boldsymbol{\nu}_{r,k}^{\top} &
\boldsymbol{a}_{k}^{\top}
\end{bmatrix}^{\top}
=
\begin{bmatrix}
u_{r,k} & v_{r,k} & r_k & n_k & \delta_k
\end{bmatrix}^{\top}.
\label{eq:relative_state_input}
\end{equation}
For a history length $L$, the recent maneuvering history is written as
\begin{equation}
\mathcal{H}_{r,k}
=
\left[
\boldsymbol{s}_{r,k-L+1},
\ldots,
\boldsymbol{s}_{r,k}
\right].
\label{eq:history_window}
\end{equation}

From the perspective of function approximation, the unknown transition mapping can be approximated using suitable families of functions. For a scalar argument \mbox{\(z\)}, representative examples include polynomial and trigonometric basis expansions, expressed as
\begin{equation}
\begin{aligned}
 p_m(z)&=\sum_{i=0}^{m}a_i z^i,\\
 t_m(z)&=b_0+\sum_{i=1}^{m}
 \left[b_i\cos(i\omega z)+c_i\sin(i\omega z)\right].
\end{aligned}
\label{eq:approximation_prescribed_examples}
\end{equation}
Polynomial terms conveniently represent smooth responses over a bounded operating range, including low-order relationships suggested by mechanics. Trigonometric terms are natural for periodic responses, with \mbox{\(\omega=2\pi/T\)} for period \mbox{\(T\)}. In both examples, the basis functions are prescribed and their combination coefficients are fitted \citep{trefethen2013}.

Unlike prescribed basis expansions, neural networks can construct data-adaptive approximation functions through trainable nonlinear transformations.  For a vector input \mbox{\(\boldsymbol z\)}, a single-hidden-layer example is
\begin{equation}
\begin{aligned}
 \psi_j(\boldsymbol z)&=\sigma(\boldsymbol W_j\boldsymbol z+\beta_j),\\
 \widehat f_{\mathrm N}(\boldsymbol z)
 &=\alpha_0+\sum_{j=1}^{q}\alpha_j\psi_j(\boldsymbol z).
\end{aligned}
\label{eq:approximation_adaptive_basis}
\end{equation}
Here, the prescribed activation \mbox{\(\sigma\)} generates functions through affine transformations. Updating the hidden-layer weights \mbox{\(\boldsymbol W_j\)} and biases \mbox{\(\beta_j\)} changes \mbox{\(\psi_j\)} as functions of the input. Training therefore learns both these functions and their output combination coefficients \mbox{\(\alpha_j\)} \citep{goodfellow2016}. This gives a basis-function interpretation of neural approximation \citep{cybenko1989,hornik1991}. Deep and recurrent networks extend this view through compositions of parameterized nonlinear mappings; in an LSTM, the resulting functions depend on the input history. Here, data-adaptive basis refers to these learned approximation functions, without requiring linear independence.

Following this function-approximation view, the central problem is to construct a transition predictor that effectively leverages the complementary strengths of different basis functions to better characterize the underlying dynamics of the system being modeled. In the present study, the first component is a reduced polynomial maneuvering prior
$\boldsymbol{p}_{\boldsymbol{\theta}}(\cdot)$, whose prescribed basis functions provide a structured approximation of the dominant hydrodynamic response. The second component is a data-adaptive basis module
$\boldsymbol{R}_{\boldsymbol{\phi}}(\cdot)$, which captures nonlinear and history-dependent effects that are not adequately represented by the polynomial prior.

The objective is to learn a discrete-time maneuvering model that predicts the one-step-ahead relative velocity:
\begin{equation}
\widehat{\boldsymbol{\nu}}_{r,k+1}
=
\mathcal{C}
\left(
\boldsymbol{p}_{\boldsymbol{\theta}},
\boldsymbol{R}_{\boldsymbol{\phi}},
\mathcal{H}_{r,k}
\right),
\label{eq:problem_joint_transition}
\end{equation}
where $\mathcal{C}(\cdot)$ denotes a differentiable coupling of the polynomial prior and the data-adaptive basis module. Composite approximation refers to the joint representation of the transition using complementary function approximators, while differentiability allows their parameters to be jointly calibrated through gradient-based optimization. Its concrete residual-coupled realization is defined in Section~3.

The joint calibration problem is first written in terms of the effective polynomial coefficients as
\begin{equation}
(\boldsymbol{\theta}^{\ast},\boldsymbol{\phi}^{\ast})
=
\arg\min_{\boldsymbol{\theta},\boldsymbol{\phi}}
\sum_{i=1}^{N}
\sum_{k=L}^{T_i-1}
\ell
\left(
\boldsymbol{\nu}^{(i)}_{r,k+1},
\widehat{\boldsymbol{\nu}}^{(i)}_{r,k+1},
\Delta\boldsymbol{\nu}^{(i)}_{k+1}
\right).
\label{eq:joint_learning_objective}
\end{equation}
Here, $\Delta\boldsymbol{\nu}^{(i)}_{k+1}$ denotes the adaptive correction term, and the loss function $\ell(\cdot)$ penalizes the one-step prediction error and the amplitude of the adaptive correction. Through this objective, the polynomial coefficients and data-adaptive parameters are jointly calibrated under a shared objective that balances prediction accuracy and correction magnitude. Fixing the polynomial coefficients gives the corresponding frozen-prior calibration problem. In the implementation, \mbox{$\boldsymbol{\theta}(\boldsymbol w)$} is generated from bounded correction variables \mbox{$\boldsymbol w$}; the concrete coupling, network architecture, and calibration procedure are presented in Section~3.

\section{Methodology}

\subsection{Overview of the proposed model}

\begin{figure}[htbp]
  \centering
  \includegraphics[width=0.99\linewidth]{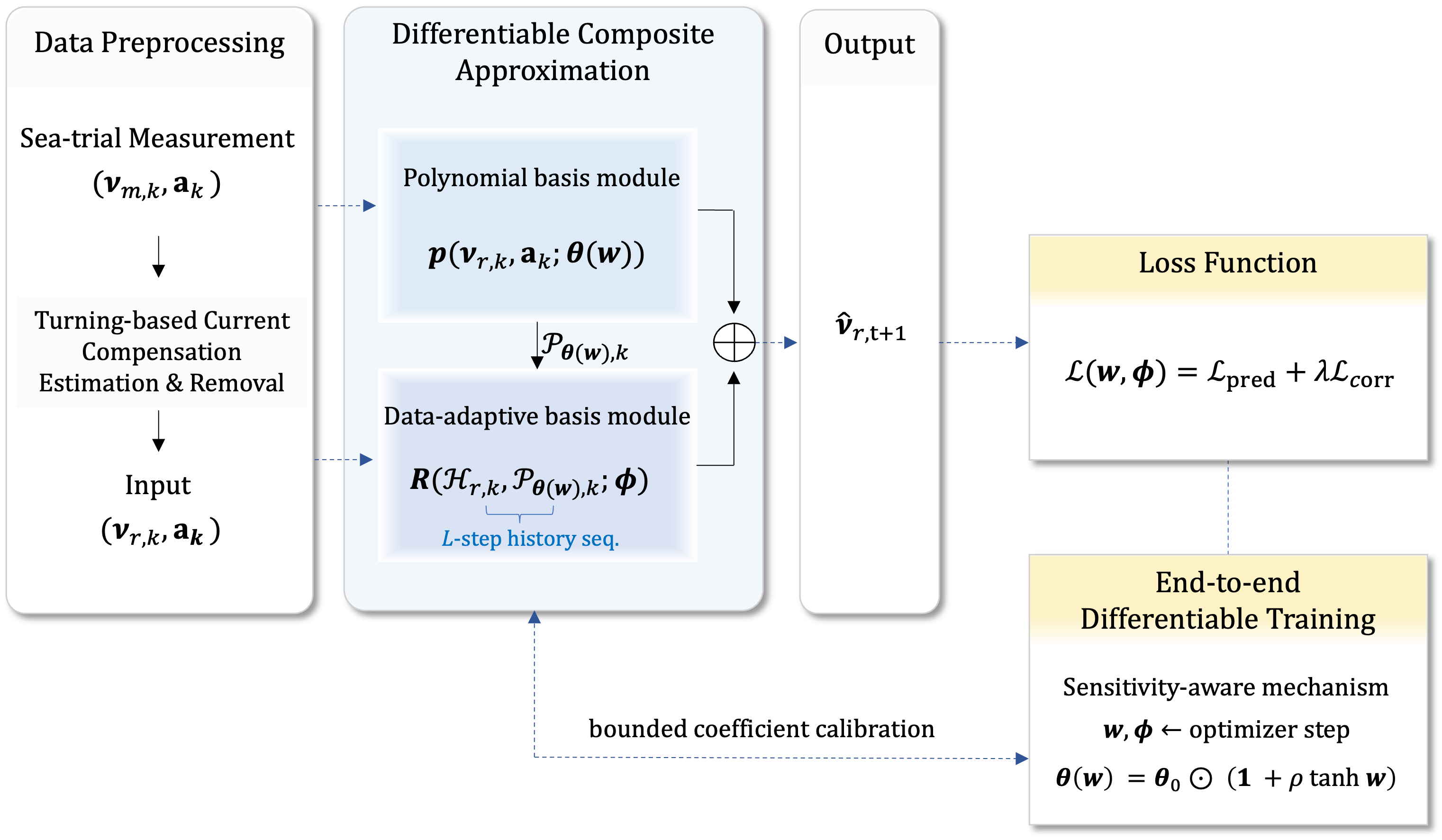}
  \caption{Proposed current-compensated differentiable composite approximation framework for AUV maneuvering modeling}
  \label{fig:modeling_framework}
\end{figure}

The proposed framework integrates a polynomial hydrodynamic model and a data-adaptive basis module into a differentiable composite approximation model. The parameters of both components are jointly calibrated by gradient descent under a common prediction loss, with a sensitivity-aware mechanism regulating the bounded updates of the polynomial component. As shown in Fig.~\ref{fig:modeling_framework}, the workflow first converts the measured sea-trial velocities into water-relative velocities by estimating and removing the current-induced component. The maneuvering dynamics are then learned in the water-relative frame using the jointly calibrated composite predictor. During rollout, the estimated current is reintroduced to reconstruct the measured field behavior.

To implement the abstract predictor in Eq.~\eqref{eq:problem_joint_transition}, the polynomial prior is evaluated over the same history window:
\begin{equation}
\mathcal{P}_{\boldsymbol{\theta}(\boldsymbol{w}),k}
=
\left[
\boldsymbol{p}_{\boldsymbol{\theta}(\boldsymbol{w})}(\boldsymbol{s}_{r,k-L+1}),
\ldots,
\boldsymbol{p}_{\boldsymbol{\theta}(\boldsymbol{w})}(\boldsymbol{s}_{r,k})
\right].
\label{eq:prior_history}
\end{equation}
The prior-output history provides the adaptive module with information about how the polynomial model explains the recent maneuvering response. The implemented residual-coupled prediction model is
\begin{equation}
\widehat{\boldsymbol{\nu}}_{r,k+1}
=
\boldsymbol{p}_{\boldsymbol{\theta}(\boldsymbol{w})}(\boldsymbol{s}_{r,k})
+
\boldsymbol{R}_{\boldsymbol{\phi}}
\left(
\mathcal{H}_{r,k},
\mathcal{P}_{\boldsymbol{\theta}(\boldsymbol{w}),k}
\right),
\label{eq:method_transition}
\end{equation}
where $\mathcal{P}_{\boldsymbol{\theta}(\boldsymbol{w}),k}$ is the prior-output history defined in Eq.~\eqref{eq:prior_history}. The polynomial module and the data-adaptive basis module are coupled through a skip connection. Specifically, the current polynomial output is directly retained as the hydrodynamic baseline in Eq.~\eqref{eq:method_transition}, while the polynomial-output history is also provided as part of the data-adaptive basis input. This skip-connected design allows the data-adaptive basis module to learn deviations conditioned on the polynomial hydrodynamic prediction, while the effective polynomial coefficients remain trainable.

\subsection{Current compensation for sea-trial data}
One major challenge in field modeling of AUVs is the disturbance induced by ocean currents. A feasible solution is to estimate the current velocity and compensate for its effect, thereby obtaining the water-relative motion states of the vehicle. This approach follows the relative-velocity formulation for marine craft under ocean currents \citep{fossen2021}.

For each short maneuvering trial, the ocean current is approximated as locally constant in the inertial frame. 
Under this approximation, the measured body-fixed velocity is expressed as the sum of the water-relative velocity and the current projected into the body frame:
\begin{equation}
\boldsymbol{\nu}^b_m
=
\boldsymbol{\nu}^b_r
+
\mathbf{R}^{\top}(\psi)\boldsymbol{\nu}^n_c .
\label{eq:7}
\end{equation}
where $\boldsymbol{\nu}^n_c=[V_c\cos\beta_c,V_c\sin\beta_c,0]^{\top}$ is the current velocity expressed in the inertial frame, and $\mathbf{R}(\psi)$ is the horizontal-plane rotation matrix from the body-fixed frame to the inertial frame.

The current speed $V_c$ and direction $\beta_c$ are estimated from the steady portion of turning maneuvers. During this segment, $(u_r,v_r,r)$ are treated as approximately constant, while the measured surge and sway velocities vary periodically with the heading angle because the inertial current is observed in the rotating body frame. The resulting regression model is
\begin{equation}
\boldsymbol{\nu}^b_m
=
\boldsymbol{\Phi}\boldsymbol{\gamma}
=
\begin{bmatrix}
1 & 0 & 0 & \cos\psi & \sin\psi \\
0 & 1 & 0 & -\sin\psi & \cos\psi \\
0 & 0 & 1 & 0 & 0
\end{bmatrix}
\begin{bmatrix}
u_r \\
v_r \\
r \\
V_c\cos(\beta_c) \\
V_c\sin(\beta_c)
\end{bmatrix}.
\label{eq:8}
\end{equation}
Stacking $M$ samples from the steady turning segment gives
\begin{equation}
\boldsymbol{Y}
=
\begin{bmatrix}
\boldsymbol{\nu}^b_m(1)\\
\boldsymbol{\nu}^b_m(2)\\
\vdots\\
\boldsymbol{\nu}^b_m(M)
\end{bmatrix},
\qquad
\boldsymbol{H}
=
\begin{bmatrix}
\boldsymbol{\Phi}(1)\\
\boldsymbol{\Phi}(2)\\
\vdots\\
\boldsymbol{\Phi}(M)
\end{bmatrix}.
\label{eq:9}
\end{equation}
The unknown parameter vector is estimated by least squares:
\begin{equation}
\hat{\boldsymbol{\gamma}}
=
\arg\min_{\boldsymbol{\gamma}}
\left\|
\boldsymbol{Y}
-
\boldsymbol{H}\boldsymbol{\gamma}
\right\|_2^2 .
\label{eq:10}
\end{equation}
The continuous change of $\psi$ in a turning maneuver provides the trigonometric excitation needed for the current components in Eq.~\eqref{eq:8}. With $\hat{\gamma}_4$ and $\hat{\gamma}_5$ denoting the estimated components of $V_c\cos\beta_c$ and $V_c\sin\beta_c$, respectively, the current magnitude and direction are reconstructed as
\begin{equation}
\begin{aligned}
\hat{V}_c
&=
\sqrt{
\hat{\gamma}_4^2
+
\hat{\gamma}_5^2
},\\
\hat{\beta}_c
&=
\operatorname{atan2}
\left(
\hat{\gamma}_5,
\hat{\gamma}_4
\right).
\end{aligned}
\label{eq:11}
\end{equation}
For notational simplicity, the hats are omitted in the subsequent current-compensation and rollout equations.

After outlier replacement and low-pass filtering of the measured signals, the current-compensated velocity sequence used for training is constructed by removing the estimated body-frame current component:
\begin{equation}
\boldsymbol{\nu}^b_r
=
\boldsymbol{\nu}^b_m
-
\begin{bmatrix}
V_c\cos(\beta_c-\psi)\\
V_c\sin(\beta_c-\psi)\\
0
\end{bmatrix}.
\label{eq:17}
\end{equation}
Based on the current-compensated velocity sequence, the maneuvering dynamics can be learned in the water-relative frame.

\subsection{Differentiable composite approximation architecture}
In this study, the maneuvering model is constructed as a differentiable composite approximation of the one-step maneuvering transition. It contains two complementary approximation modules. The first module uses explicitly specified polynomial hydrodynamic basis functions, which provide a compact and interpretable representation of the dominant surge, sway, and yaw-rate responses. The second module is a data-adaptive basis module implemented using long short-term memory (LSTM) networks, which learn history-dependent nonlinear features from the maneuvering sequence. The composite predictor is calibrated as a single differentiable transition map.

For $\boldsymbol{s}_{r,k}=[u_{r,k},v_{r,k},r_k,n_k,\delta_k]^{\top}$, the polynomial basis module is written as
\begin{equation}
\boldsymbol{p}_{\boldsymbol{\theta}}(\boldsymbol{s}_{r,k})
=
\begin{bmatrix}
p_u(\boldsymbol{s}_{r,k})\\
p_v(\boldsymbol{s}_{r,k})\\
p_r(\boldsymbol{s}_{r,k})
\end{bmatrix}.
\label{eq:polynomial_prior_vector}
\end{equation}
where $\boldsymbol{p}_{\boldsymbol{\theta}}=[p_u,p_v,p_r]^{\top}$ directly approximates the next-step water-relative velocity. The retained channel-wise basis functions are
\begin{equation}
\begin{aligned}
p_u
&=
\theta_{u,1}u_r
+
\theta_{u,2}u_r^2
+
\theta_{u,3}n
+
\theta_{u,4}n^2
+
\theta_{u,5}\delta^2u_r^2,\\
p_v
&=
\theta_{v,1}v_r
+
\theta_{v,2}r
+
\theta_{v,3}\delta
+
\theta_{v,4}r^3
+
\theta_{v,5}\delta^2 r,\\
p_r
&=
\theta_{r,1}r
+
\theta_{r,2}r^3
+
\theta_{r,3}\delta
+
\theta_{r,4}u_r r
+
\theta_{r,5}r\delta^2
+
\theta_{r,6}v_r r^2 .
\end{aligned}
\label{eq:polynomial_prior_basis}
\end{equation}
This module is a reduced one-step transition approximation rather than a full force–moment model or a complete Abkowitz-type third-order polynomial maneuvering model. 
The retained basis includes the main linear damping terms and selected nonlinear terms, allowing the module to capture the dominant hydrodynamic response with a compact parameterization. The corresponding coefficients are therefore interpreted as effective coefficients of the retained polynomial transition basis. In this sense, the module serves as an imperfect mechanistic prior that remains trainable during joint calibration with the data-adaptive basis module, while avoiding the parameter drift and generalization issues that may arise when estimating a fully polynomial model from limited sea-trial data.

Regarding the data-adaptive basis module, LSTM networks are used because the correction to the one-step transition may depend on recent maneuvering memory, especially after steering actions and during the transient development of sway and yaw motion. In this study, the LSTM is treated as a standard recurrent nonlinear mapping. Its internal cell state allows the final hidden state to encode the recent maneuvering history. The initial hidden and cell states are set to zero for each training window.

For each history window, the polynomial module is evaluated at every time step using the unnormalized physical variables in $\mathcal{H}_{r,k}$, producing the prior-output history $\mathcal{P}_{\boldsymbol{\theta}(\boldsymbol{w}),k}$ in Eq.~\eqref{eq:prior_history}. This sequence is then standardized using the same velocity scaling as $(u_r,v_r,r)$ and supplied to the LSTM module together with the standardized state-input variables. Let the tilde denote standardized variables. For each time index $j$ in the window, the surge-channel data-adaptive basis input is
\begin{equation}
\boldsymbol{z}^{u}_{j}
=
\begin{bmatrix}
\tilde{u}_{r,j} &
\tilde{n}_{j} &
\tilde{\delta}_{j} &
\tilde{p}_{u,j}
\end{bmatrix}^{\top},
\label{eq:surge_lstm_input}
\end{equation}
and the sway-yaw data-adaptive basis input is
\begin{equation}
\boldsymbol{z}^{vr}_{j}
=
\begin{bmatrix}
\tilde{v}_{r,j} &
\tilde{r}_{j} &
\tilde{n}_{j} &
\tilde{\delta}_{j} &
\tilde{p}_{v,j} &
\tilde{p}_{r,j}
\end{bmatrix}^{\top}.
\label{eq:sway_yaw_lstm_input}
\end{equation}

In the implemented architecture, the surge channel uses one LSTM branch and the coupled sway-yaw channels use another LSTM branch. The final hidden states are mapped by fully connected output layers to adaptive correction terms:
\begin{equation}
\Delta u_{k+1}
=
\boldsymbol{W}^{u}_{y}\boldsymbol{h}^{u}_{k}
+
\boldsymbol{\beta}^{u}_{y},
\qquad
\boldsymbol{h}^{u}_{k}
=
\operatorname{LSTM}_{u,\boldsymbol{\phi}_u}
\left(
\boldsymbol{z}^{u}_{k-L+1:k}
\right),
\label{eq:surge_residual_branch}
\end{equation}
and
\begin{equation}
\begin{bmatrix}
\Delta v_{k+1}\\
\Delta r_{k+1}
\end{bmatrix}
=
\boldsymbol{W}^{vr}_{y}\boldsymbol{h}^{vr}_{k}
+
\boldsymbol{\beta}^{vr}_{y},
\qquad
\boldsymbol{h}^{vr}_{k}
=
\operatorname{LSTM}_{vr,\boldsymbol{\phi}_{vr}}
\left(
\boldsymbol{z}^{vr}_{k-L+1:k}
\right).
\label{eq:sway_yaw_residual_branch}
\end{equation}
The LSTM parameters and the output-layer parameters are included in $\boldsymbol{\phi}_u$ and $\boldsymbol{\phi}_{vr}$.
The data-adaptive basis output is collected as
\begin{equation}
\Delta\boldsymbol{\nu}_{k+1}
=
\begin{bmatrix}
\Delta u_{k+1} &
\Delta v_{k+1} &
\Delta r_{k+1}
\end{bmatrix}^{\top}.
\label{eq:residual_vector}
\end{equation}
Equivalently, the compact notation used in Eq.~\eqref{eq:method_transition} is
\begin{equation}
\boldsymbol{R}_{\boldsymbol{\phi}}
\left(
\mathcal{H}_{r,k},
\mathcal{P}_{\boldsymbol{\theta}(\boldsymbol{w}),k}
\right)
=
\Delta\boldsymbol{\nu}_{k+1}.
\label{eq:adaptive_basis_compact}
\end{equation}
The final prediction uses a skip connection from the polynomial module to the output:
\begin{equation}
\widehat{\boldsymbol{\nu}}_{r,k+1}
=
\boldsymbol{p}_{\boldsymbol{\theta}(\boldsymbol{w})}(\boldsymbol{s}_{r,k})
+
\Delta\boldsymbol{\nu}_{k+1}.
\label{eq:prior_residual_sum}
\end{equation}
This skip-connected structure has two roles. First, the current polynomial output is directly retained as the structured one-step baseline. Second, the polynomial-output history is provided to the LSTM module as part of its sequence input, so that the data-adaptive basis is conditioned on the hydrodynamic prediction over the recent history. Thus, the polynomial basis module and the data-adaptive basis module form a serial--parallel composite approximation that remains differentiable with respect to both the polynomial coefficients and the LSTM parameters.

\subsection{Joint optimization and sensitivity-aware bounded calibration}
After defining the composite approximation architecture, the remaining task is to calibrate the polynomial basis module and the data-adaptive basis module in a single optimization problem. The procedure contains three parts. First, an initial polynomial transition model is obtained from current-compensated data to provide a structured starting point. Second, the polynomial coefficients are converted into bounded trainable variables so that the polynomial module can be adjusted during learning without drifting arbitrarily. Third, the bounded coefficient variables and the data-adaptive basis parameters are optimized by the same one-step prediction loss. This forms a sensitivity-aware calibration mechanism because the allowable coefficient movement and the gradient sensitivity of each coefficient are explicitly regulated by the bounded parameterization.

The initial polynomial coefficients are obtained by fitting the retained basis functions to the current-compensated training data:
\begin{equation}
\boldsymbol{\theta}_0
=
\arg\min_{\bar{\boldsymbol{\theta}}}
\sum_{i=1}^{N}
\sum_{k=1}^{T_i-1}
\left\|
\boldsymbol{\nu}^{(i)}_{r,k+1}
-
\boldsymbol{p}_{\bar{\boldsymbol{\theta}}}
(\boldsymbol{s}^{(i)}_{r,k})
\right\|_2^2 .
\label{eq:initial_prior_fit}
\end{equation}
This initial fit provides a structured reference model, but it is not treated as the final fixed physical model. In the proposed method, the effective coefficients used by the polynomial module are generated from bounded correction variables:
\begin{equation}
\boldsymbol{\theta}(\boldsymbol{w})
=
\boldsymbol{\theta}_0
\odot
\left(
\boldsymbol{1}
+
\rho \tanh\boldsymbol{w}
\right),
\label{eq:bounded_coefficient_calibration}
\end{equation}
where $\boldsymbol{w}$ is trainable, $\rho$ is the prescribed maximum relative correction range, and $\odot$ denotes element-wise multiplication. This parameterization gives
\begin{equation}
-\rho
<
\frac{\theta_j-\theta_{0,j}}{\theta_{0,j}}
<
\rho ,
\label{eq:coefficient_relative_bound}
\end{equation}
for nonzero $\theta_{0,j}$. Thus, the polynomial basis module remains trainable, but its effective coefficients are constrained to stay within a controlled neighborhood of the initial polynomial fit.

The trainable variables are the coefficient-correction variables $\boldsymbol{w}$ and the data-adaptive basis parameters $\boldsymbol{\phi}=\{\boldsymbol{\phi}_u,\boldsymbol{\phi}_{vr}\}$. Let $q=1,\ldots,N_s$ index the valid training windows, and let $k_q$ denote the last time index in the $q$th window. The one-step prediction error for this window is
\begin{equation}
\boldsymbol{e}_{q}
=
\boldsymbol{\nu}_{r,k_q+1}
-
\widehat{\boldsymbol{\nu}}_{r,k_q+1}.
\label{eq:one_step_error}
\end{equation}
The training objective is
\begin{equation}
\mathcal{L}(\boldsymbol{w},\boldsymbol{\phi})
=
\frac{1}{N_s}
\sum_{q=1}^{N_s}
\left[
\boldsymbol{e}_{q}^{\top}
\boldsymbol{W}_{\nu}
\boldsymbol{e}_{q}
+
\lambda
\Delta\boldsymbol{\nu}_{q}^{\top}
\boldsymbol{D}_{\nu}^{-2}
\Delta\boldsymbol{\nu}_{q}
\right],
\label{eq:method_training_loss}
\end{equation}
where $\Delta\boldsymbol{\nu}_{q}$ is the adaptive correction generated by $\boldsymbol{R}_{\boldsymbol{\phi}}$ for the same window, $\boldsymbol{D}_{\nu}=\operatorname{diag}(\sigma_u,\sigma_v,\sigma_r)$ contains the training-set standard deviations of the velocity channels, $\boldsymbol{W}_{\nu}=\operatorname{diag}(\omega_u,\omega_v,\omega_r)$ is a channel-weighting matrix with $\omega_{\ell}$ proportional to $1/\sigma_{\ell}^{2}$ and normalized by its mean, and $\lambda$ is the adaptive-correction penalty coefficient. The first term fits the next water-relative velocity, while the second term discourages the data-adaptive basis module from unnecessarily replacing the polynomial basis module.

The joint training problem can now be written as
\begin{equation}
(\boldsymbol{w}^{\ast},\boldsymbol{\phi}^{\ast})
=
\arg\min_{\boldsymbol{w},\boldsymbol{\phi}}
\mathcal{L}(\boldsymbol{w},\boldsymbol{\phi}).
\label{eq:method_joint_optimization}
\end{equation}
In optimization, $\boldsymbol{w}$ and $\boldsymbol{\phi}$ are included in the same computational graph and updated by a gradient-based optimizer with separate parameter groups and learning rates.

The coefficient update is obtained by back-propagation through the differentiable composite predictor. For an individual correction variable $w_j$,
\begin{equation}
\frac{\partial \theta_j}{\partial w_j}
=
\rho\theta_{0,j}
\left(
1-\tanh^2 w_j
\right),
\label{eq:bounded_gradient}
\end{equation}
Equation~\eqref{eq:bounded_gradient} shows the sensitivity-aware nature of the coefficient calibration. The scalar $\rho$ sets the maximum relative range, while the factor $1-\tanh^2 w_j$ reduces the local coefficient sensitivity as the correction approaches its bound. In one-step training, the history window is constructed from measured current-compensated samples. Under this training setting, the sensitivity of the prediction to $w_j$ contains both the direct polynomial-baseline path and the indirect data-adaptive basis input path:
\begin{equation}
\frac{\partial \widehat{\boldsymbol{\nu}}_{r,k+1}}
{\partial w_j}
=
\left[
\frac{\partial \boldsymbol{p}_{\boldsymbol{\theta}(\boldsymbol{w})}(\boldsymbol{s}_{r,k})}
{\partial \theta_j}
+
\frac{\partial \boldsymbol{R}_{\boldsymbol{\phi}}}
{\partial \mathcal{P}_{\boldsymbol{\theta}(\boldsymbol{w}),k}}
\frac{\partial \mathcal{P}_{\boldsymbol{\theta}(\boldsymbol{w}),k}}
{\partial \theta_j}
\right]
\frac{\partial \theta_j}{\partial w_j}.
\label{eq:cooptimization_gradient}
\end{equation}
Here, the two terms correspond to the direct and indirect gradient paths from the polynomial module to the final prediction. The direct path comes from the skip connection, where the last polynomial output in the input window is added to the prediction, while the indirect path comes from feeding the standardized polynomial-output history into the data-adaptive basis module. In addition, the adaptive-correction penalty in Eq.~\eqref{eq:method_training_loss} also contributes a differentiable gradient through $\Delta\boldsymbol{\nu}_{q}$.

Algorithm~\ref{alg:joint_training} summarizes the training procedure. 

\begin{algorithm}[htbp]
\caption{Joint training of the polynomial basis and data-adaptive basis modules}
\label{alg:joint_training}
\KwIn{Current-compensated training trials; history length $L$; initial polynomial coefficients $\boldsymbol{\theta}_0$; correction range $\rho$; adaptive-correction penalty $\lambda$; learning rates $\eta_{\phi}$ and $\eta_{w}$.}
\KwOut{Optimized effective polynomial coefficients $\boldsymbol{\theta}(\boldsymbol{w}^{\ast})$ and data-adaptive basis parameters $\boldsymbol{\phi}^{\ast}$.}
Compute training-set normalization statistics from the current-compensated training data\;
Construct unstandardized windows $\mathcal{H}_{r,k_q}$, standardized windows $\widetilde{\mathcal{H}}_{r,k_q}$, and targets $\boldsymbol{\nu}_{r,k_q+1}$\;
Initialize $\boldsymbol{w}=\boldsymbol{0}$ and initialize $\boldsymbol{\phi}$\;
Create optimizer parameter groups for $\boldsymbol{\phi}$ with learning rate $\eta_{\phi}$ and for $\boldsymbol{w}$ with learning rate $\eta_{w}$ and no weight decay\;
\For{each training epoch}{
  \For{each mini-batch $\mathcal{B}$}{
    Compute bounded effective polynomial coefficients
    $\boldsymbol{\theta}(\boldsymbol{w})=\boldsymbol{\theta}_0\odot(\boldsymbol{1}+\rho\tanh\boldsymbol{w})$\;
    \For{each window $q\in\mathcal{B}$}{
      Evaluate $\boldsymbol{p}_{\boldsymbol{\theta}(\boldsymbol{w})}(\boldsymbol{s}_{r,j})$ for $j=k_q-L+1,\ldots,k_q$ to form $\mathcal{P}_{\boldsymbol{\theta}(\boldsymbol{w}),k_q}$\;
      Standardize $\mathcal{P}_{\boldsymbol{\theta}(\boldsymbol{w}),k_q}$ and build $\boldsymbol{z}^{u}_{k_q-L+1:k_q}$ and $\boldsymbol{z}^{vr}_{k_q-L+1:k_q}$ from Eqs.~\eqref{eq:surge_lstm_input} and~\eqref{eq:sway_yaw_lstm_input}\;
      Compute the adaptive correction
      $\Delta\boldsymbol{\nu}_{q}
      =
      \boldsymbol{R}_{\boldsymbol{\phi}}
      (\mathcal{H}_{r,k_q},\mathcal{P}_{\boldsymbol{\theta}(\boldsymbol{w}),k_q})$\;
      Form the one-step prediction
      $\widehat{\boldsymbol{\nu}}_{r,k_q+1}
      =
      \boldsymbol{p}_{\boldsymbol{\theta}(\boldsymbol{w})}(\boldsymbol{s}_{r,k_q})
      +
      \Delta\boldsymbol{\nu}_{q}$\;
    }
    Compute the mini-batch version of the loss in Eq.~\eqref{eq:method_training_loss}\;
    Clear accumulated gradients and back-propagate the loss through $\boldsymbol{\theta}(\boldsymbol{w})$, $\mathcal{P}_{\boldsymbol{\theta}(\boldsymbol{w}),k_q}$, the data-adaptive basis branches, and the skip-connected summation\;
    Update the optimizer parameter groups for $\boldsymbol{\phi}$ and $\boldsymbol{w}$\;
  }
}
\end{algorithm}

\subsection{Rollout prediction and current reintroduction}
During model application, the learned predictor can be used in two modes. In the first mode, it is executed directly in the water-relative frame as an inherent maneuvering model to evaluate the vehicle maneuverability. In the second mode, the estimated current is reintroduced during rollout to predict the current-affected vehicle motion observed in sea trials. 

For both modes, the model is executed over complete maneuvering trials. The measured initial history is used to initialize the input window; after each step, the predicted water-relative velocity is appended to the history together with the recorded exogenous inputs $(n,\delta)$ for the next prediction. This protocol tests the accumulated error behavior relevant to maneuvering simulation, digital-twin construction, and model-based control.

For the current-reintroduced mode, the predicted yaw rate is first integrated to update the vehicle heading. The estimated current is then projected into the body-fixed frame using the updated heading, and the measured velocity is reconstructed as
\begin{equation}
\widehat{\boldsymbol{\nu}}^b_{m,k}
=
\widehat{\boldsymbol{\nu}}^b_{r,k}
+
\begin{bmatrix}
V_c\cos(\beta_c-\widehat{\psi}_k)\\
V_c\sin(\beta_c-\widehat{\psi}_k)\\
0
\end{bmatrix}.
\label{eq:25}
\end{equation}
The reconstructed body-frame velocity is then integrated through the horizontal-plane kinematics to obtain the predicted trajectory.

\section{Experimental Validation and Analysis}

\subsection{Experimental Setup}
\begin{figure}[h]
  \centering
  \includegraphics[width=0.7\linewidth]{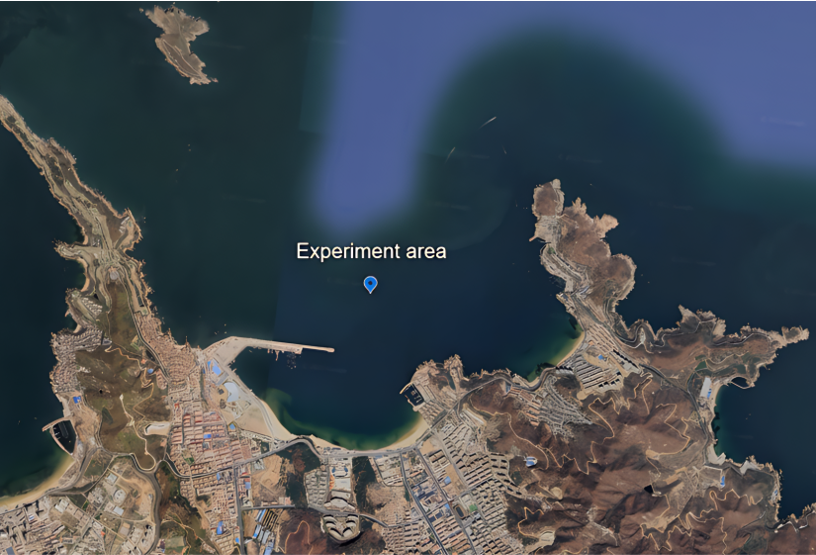}
  \caption{Geographical location and surrounding terrain of the experimental area.}
  \label{fig:experiment area}
\end{figure}

To evaluate the effectiveness of the proposed modeling method, in-field experiments were conducted using a 7-meter-long AUV under real sea conditions. The trials took place in the East China Sea near Weihai, Shandong Province (approximately 37.5450°N, 122.0908°E). The geographic location and surrounding terrain of the test area are shown in Fig.~\ref{fig:experiment area}. The coastal environment featured generally calm sea states, but local ocean currents varied significantly due to seabed topography and coastal features. For example, turbulent flow was observed near protruding shoreline regions. Such spatial variability introduced uncertainties in the AUV’s motion response, influencing the accuracy of dynamic modeling.
\begin{figure}[h]
  \centering
  \includegraphics[width=0.7\linewidth]{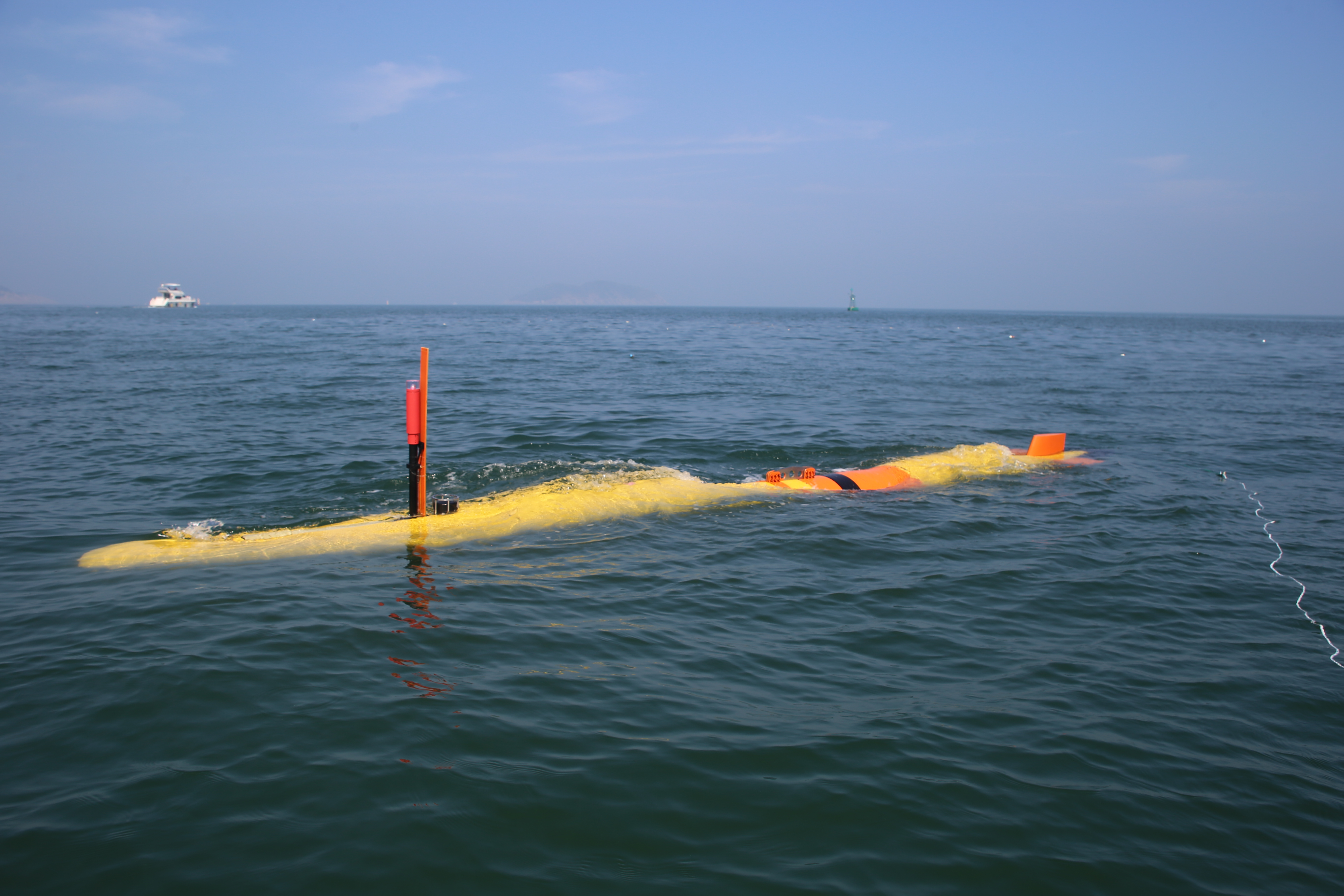}
  \caption{On-site sea trial of the experimental AUV platform.}
  \label{fig:auv}
\end{figure}

The experimental platform is a torpedo-shaped AUV measuring 7 meters in length and 0.5 meters in diameter, with a total weight of approximately 1.2 tons (as shown in Fig.~\ref{fig:auv}). It is driven by a stern-mounted screw propeller providing continuous forward thrust. Yaw and pitch control are achieved via vertical and horizontal rudders, respectively. The vertical rudders range from -23$^\circ$ to 23$^\circ$. The vehicle is equipped with an Inertial Measurement Unit (IMU) and a Doppler Velocity Log (DVL), enabling real-time acquisition of position, velocity, and attitude data. These onboard sensor measurements, along with control inputs, constitute the core dataset for dynamic modeling.

A series of free-running maneuvering trials were conducted under various operating conditions, including turning, zigzag, and random-rudder maneuvers. Each maneuver type was performed at two propeller revolutions: 280 rpm and 420 rpm, corresponding approximately to the 4 kt and 6 kt speed conditions used in the data labels. Before each trial, the AUV was submerged to a depth of 4 meters and maintained in a steady state before the commanded rudder sequence was applied. During the maneuvering trials, an independent depth controller was employed to maintain the prescribed depth by regulating the horizontal rudders. This depth-control system operated separately from the horizontal-plane maneuvering control, representing a standard operating mode of the AUV.
In the modeling phase, the maneuvering datasets serve distinct roles. Turning maneuvers are used to estimate trial-level current parameters for current compensation. For model training, the random-rudder maneuvers and selected turning maneuvers are used. The remaining turning and zigzag maneuvers are withheld from predictor training and used for validation.

\subsection{Ocean Current Estimation Results}
As detailed in the methodology section, ocean current parameters are estimated from the steady-state phase of turning maneuvers. The estimated results from different turning trials are presented in Table~\ref{tab:current_estimation}, covering propeller revolutions of 280 rpm and 420 rpm. These two sets of trials were conducted in separate sessions. In each session, the maneuvers were performed in sequence, starting with larger rudder angles and gradually decreasing to smaller ones.

\begin{table}[htbp]
  \refstepcounter{table}
  \noindent
  \makebox[\textwidth][l]{\textbf{Table \thetable }} \\ 
  \noindent
  \makebox[\textwidth][l]{Ocean current estimation from turning maneuvers} \\ 
  \centering
  \renewcommand{\arraystretch}{1.2}
  \begin{tabularx}{\textwidth}{l *{5}{>{\centering\arraybackslash}X}} 
    \toprule
    Test condition & $n$ (rpm) & $u_r$ (m/s) & $v_r$ (m/s) & $V_c$ (m/s) & $\beta_c$ (°) \\
    \midrule
    $23^{\circ}$ turning  & 280 & 1.4042 & -0.0674 & 0.2170 & 42.31 \\
    $15^{\circ}$ turning  & 280 & 1.5036 & -0.0470 & 0.2758 & 36.44 \\
    $10^{\circ}$ turning  & 280 & 1.5639 & -0.0228 & 0.3592 & 55.47 \\
    $23^{\circ}$ turning  & 420 & 2.1752 & -0.1098 & 0.3054 & 64.19 \\
    $15^{\circ}$ turning  & 420 & 2.3271 & -0.0724 & 0.3463 & 69.14 \\
    $10^{\circ}$ turning  & 420 & 2.5099 & -0.0203 & 0.6157 & 79.77 \\
    \bottomrule
  \end{tabularx} 
  \label{tab:current_estimation}
\end{table}
Across trials, the method estimates current parameters and the corresponding steady-state velocities \((u_r, v_r)\), which characterize the AUV’s turning response relative to the surrounding water. Under the 280 rpm condition, estimated current speeds $V_c$ range from 0.2170 m/s to 0.3592 m/s, with directions $\beta_c$ between 36.44$^\circ$ and 55.47$^\circ$, indicating moderate variation across trials. For the 420 rpm condition, the first two trials produce similar results, while the third shows a sharp increase to 0.6157 m/s at 79.77$^\circ$, suggesting the vehicle encountered different local current characteristics, possibly due to spatial variation in the coastal test area.

\begin{figure}[htbp]
  \centering
  \begin{subfigure}[t]{0.48\linewidth}
    \centering
    \includegraphics[width=\linewidth]{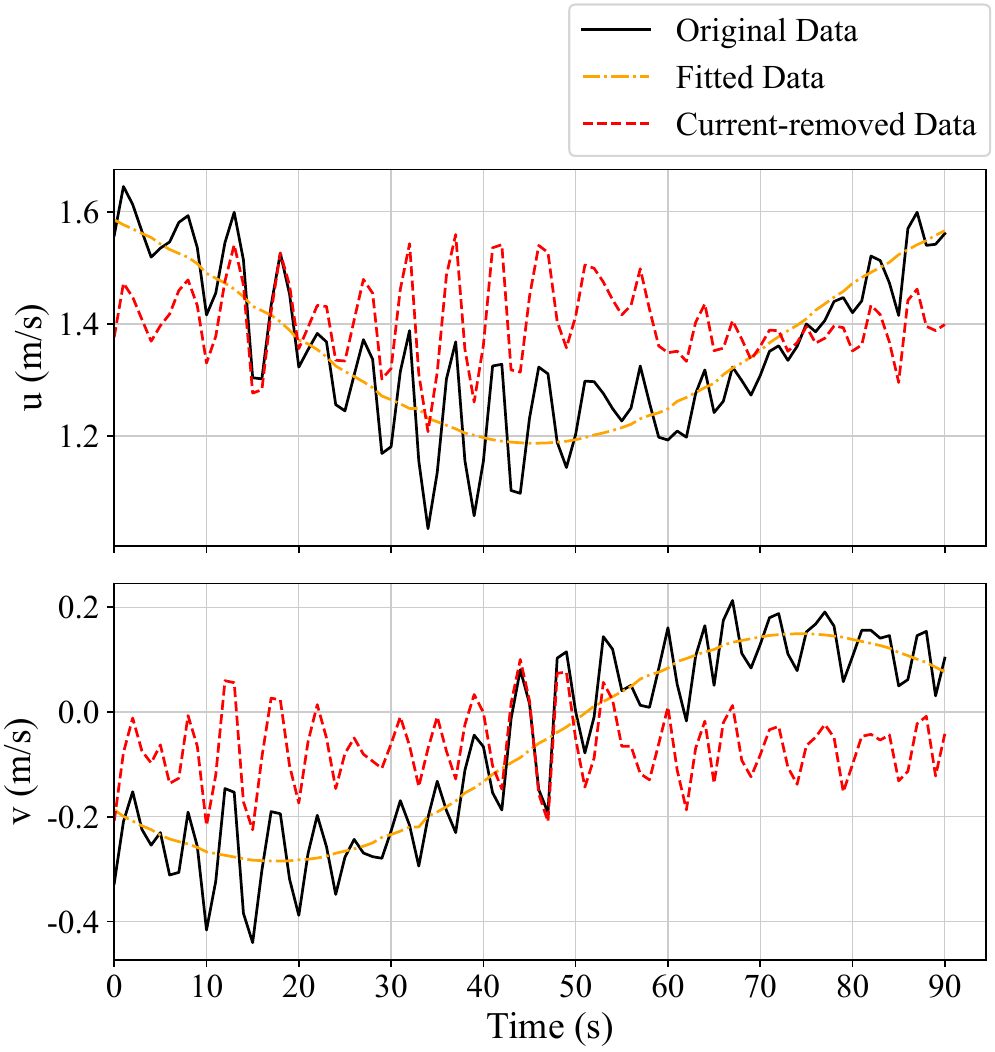}
    \caption{$23^\circ$ turning maneuver at 280 rpm.}
    \label{fig:fit1}
  \end{subfigure}
  \hfill
  \begin{subfigure}[t]{0.48\linewidth}
    \centering
    \includegraphics[width=\linewidth]{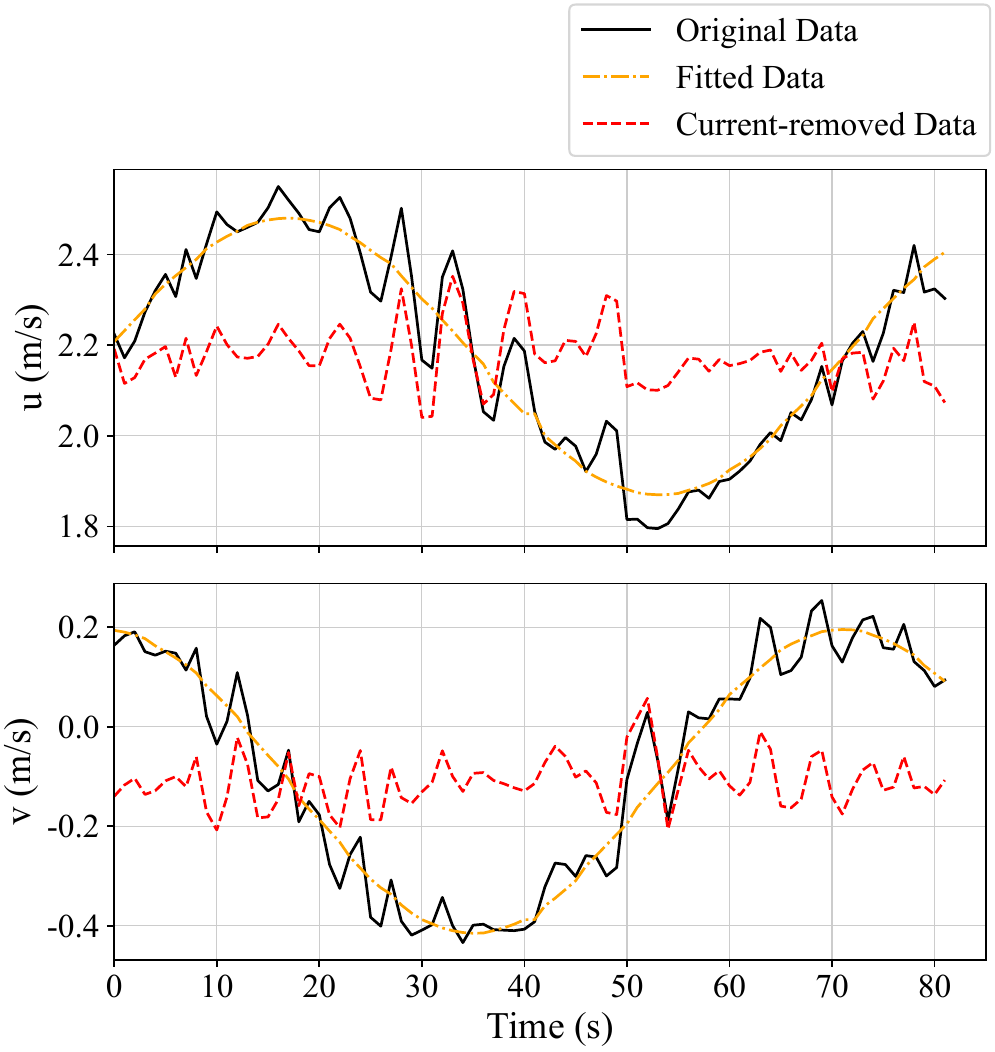}
    \caption{$23^\circ$ turning maneuver at 420 rpm.}
    \label{fig:fit2}
  \end{subfigure}
   \caption{Estimation and compensation of ocean current effects.}
  \label{fig:fit_velocities}
\end{figure}

Fig.~\ref{fig:fit_velocities} illustrates the processing of surge and sway velocities in a 23$^\circ$ turning maneuver. Under calm water conditions, these velocities are expected to approach quasi-steady values after the turn has developed. However, the raw measurements exhibit periodic variations induced by the projection of a locally constant current into the rotating body frame, overlaid with high-frequency oscillations from sensor noise and other environmental disturbances. These periodic patterns arise from the continuous change in the vehicle heading angle, which excites the trigonometric components in Eq.~\mbox{\eqref{eq:8}}. The fitted results based on Eq.~\mbox{\eqref{eq:8}} show good agreement with the measured data, capturing the periodic fluctuations attributed to the ocean current.

With the estimated current parameters, the current-induced velocity is removed using Eq.~\mbox{\eqref{eq:17}}. The compensated surge and sway velocities exhibit nearly constant trends despite minor high-frequency oscillations, which is consistent with the expected steady-state behavior during turning maneuvers. These results confirm that the current information is effectively estimated.

\subsection{Model Construction and Evaluation}
\noindent
This section evaluates whether the differentiable composite approximation can serve as an executable maneuvering model under sea-trial conditions. The evaluation is organized around three aspects. First, the proposed model is compared with a standalone polynomial prior and recurrent neural baselines to examine whether the two approximation components are complementary in recursive prediction. Second, ablation studies are conducted to separate the effects of bounded coefficient calibration and current compensation. Third, the calibrated polynomial coefficients are inspected to assess whether the prior module remains active after joint optimization. All reported errors are computed from recursive rollout prediction.

\subsubsection{Training and validation setup}
Table~\ref{tab:exp_conditions} summarizes the data used for training and validation. The random-rudder training data are compensated using the current estimate identified from the 10$^\circ$ turning maneuver at 420 rpm, which was conducted close in time to the random-rudder trials. The selected 23$^\circ$ turning trials use their corresponding same-trial current estimates. The remaining turning and zigzag maneuvers are used for generalization validation.

\begin{table}[htbp]
\centering
\caption{Training and validation data used by the proposed model}
\label{tab:exp_conditions}
{\small
\renewcommand{\arraystretch}{1.18}
\setlength{\tabcolsep}{2pt}
\begin{tabularx}{\textwidth}{>{\raggedright\arraybackslash}p{0.14\textwidth} >{\raggedright\arraybackslash}p{0.34\textwidth} X}
\toprule
Item & Maneuver data & Current estimate used \\
\midrule
\multirow{4}{*}{Training} &
4 kt random-rudder &
\multirow{2}{0.43\textwidth}{10$^\circ$ turning at 420 rpm} \\
& 6 kt random-rudder & \\
& 4 kt 23$^\circ$ turning &
\multirow{2}{0.43\textwidth}{Same-trial estimate} \\
& 6 kt 23$^\circ$ turning & \\
\addlinespace[4pt]
\midrule
\addlinespace[2pt]
\multirow{8}{*}{Validation} &
4 kt 10$^\circ$ turning &
\multirow{8}{0.43\textwidth}{Corresponding trial-level estimates in Table~\ref{tab:current_estimation} are used for reconstruction and validation comparison.} \\
& 4 kt 15$^\circ$ turning & \\
& 6 kt 10$^\circ$ turning & \\
& 6 kt 15$^\circ$ turning & \\
& 4 kt 10$^\circ$/10$^\circ$ zigzag & \\
& 4 kt 15$^\circ$/15$^\circ$ zigzag & \\
& 4 kt 20$^\circ$/20$^\circ$ zigzag & \\
& 6 kt 15$^\circ$/15$^\circ$ zigzag & \\
\bottomrule
\end{tabularx}
}
\end{table}

The proposed model uses a history length of $L=5$ with a sampling interval of 1 s. The surge data-adaptive basis branch is implemented as an LSTM with 128 hidden units followed by a fully connected output layer, while the coupled sway-yaw branch uses an LSTM with 256 hidden units followed by two fully connected output heads. The polynomial module is evaluated using the unnormalized variables $[u_r,v_r,r,n,\delta]$, whereas the LSTM branches use standardized state-input variables and standardized polynomial-output features. All neural and hybrid models are trained for 400 epochs with a batch size of 64 using Adam. The neural parameters use a learning rate of $1.0\times10^{-4}$ with a weight decay of $5.0\times10^{-4}$, while the prior-correction variables use a separate learning rate of 0.006 without weight decay. The bounded-calibration range is set to $\rho=0.9$, and the adaptive-correction penalty is set to $\lambda=0.05$. These values were specified empirically, and their influence on recursive prediction is examined in Section~\mbox{\ref{sec:calibration_ablation}}. Training is performed by minimizing the one-step loss in Eq.~\eqref{eq:method_training_loss}. During recursive validation, the predicted velocity is fed back into the next input window, and the recorded control inputs are applied at each step.

\subsubsection{Comparison with polynomial and LSTM baselines}
The comparison uses the eight holdout trials in Table~\ref{tab:exp_conditions}, including four turning maneuvers and four zigzag maneuvers. The purpose is to test whether the proposed differentiable composite model improves the executable maneuvering response.

The comparison contains the proposed model and three baselines. The first baseline, denoted as the polynomial prior, retains only the reduced polynomial transition basis. The other two baselines are neural-only LSTM predictors. LSTM-DC is trained on the current-compensated water-relative velocity $\boldsymbol{\nu}_r$, with the estimated current added back only for trajectory reconstruction. LSTM-Raw uses the same pure LSTM architecture but is trained and rolled out directly on the measured velocity $\boldsymbol{\nu}_m$.

\begin{figure}[htbp]
  \centering
  \includegraphics[width=0.98\linewidth]{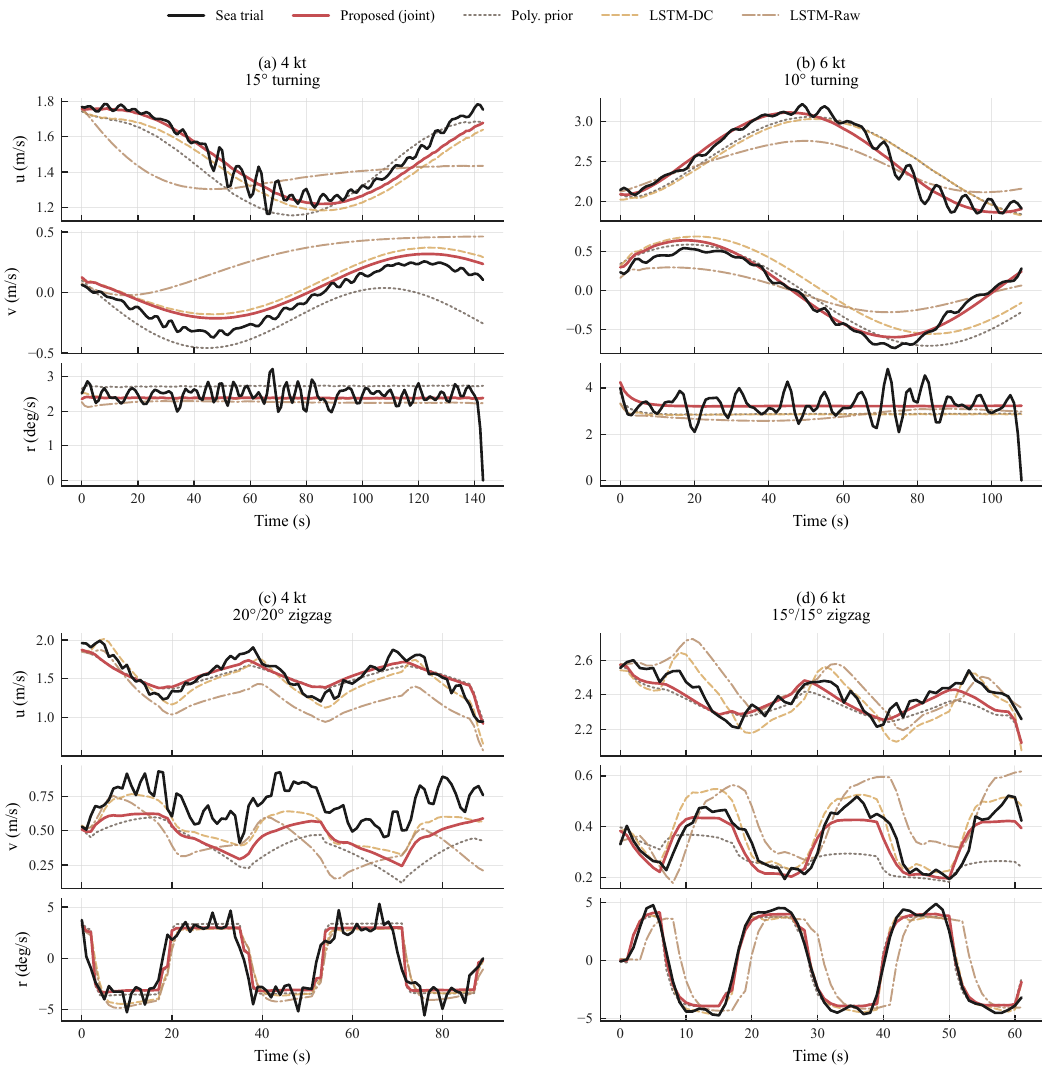}
  \caption{Representative recursive velocity predictions of the proposed model, the standalone polynomial prior, and LSTM baselines: (a) 4 kt, 15$^\circ$ turning; (b) 6 kt, 10$^\circ$ turning; (c) 4 kt, 20$^\circ$/20$^\circ$ zigzag; (d) 6 kt, 15$^\circ$/15$^\circ$ zigzag.}
  \label{fig:section43_main_representative_velocity}
\end{figure}

\begin{figure}[htbp]
  \centering
  \includegraphics[width=0.98\linewidth]{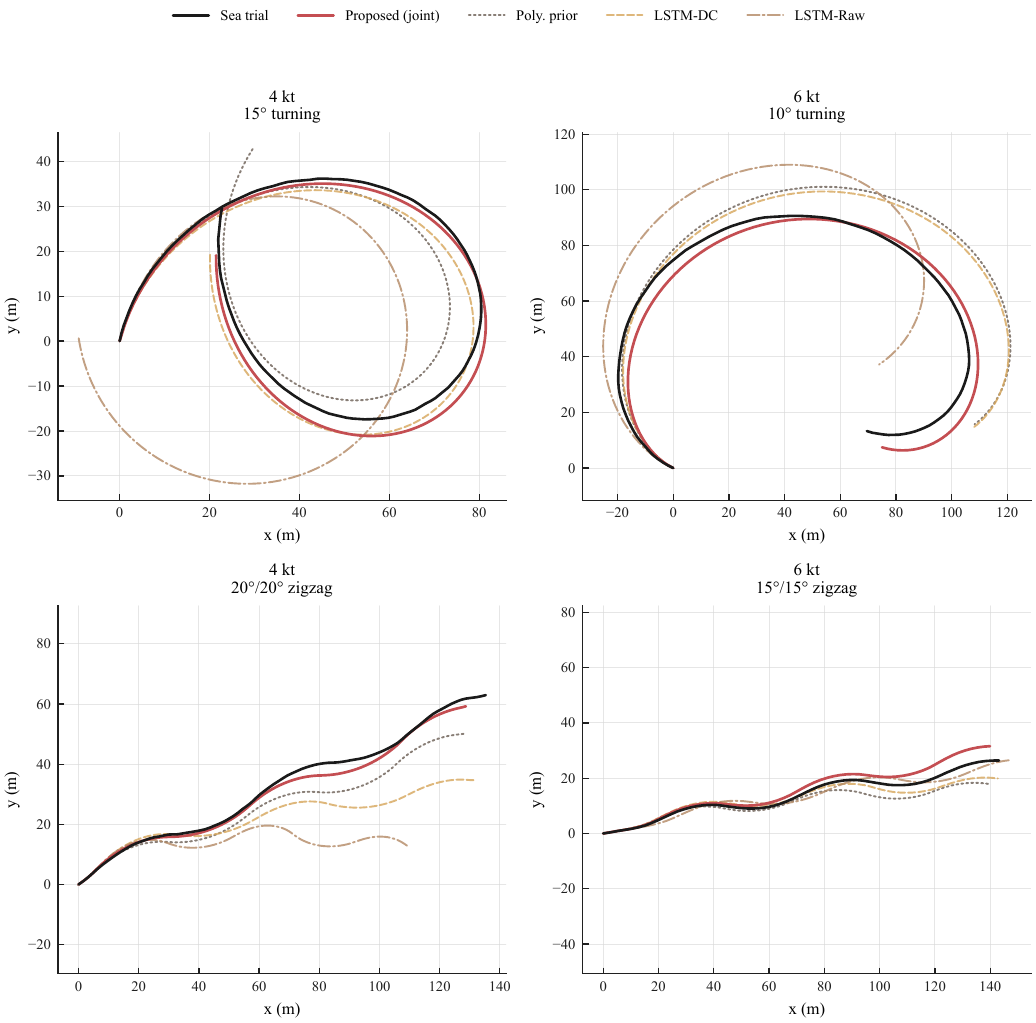}
  \caption{Representative recursive trajectory predictions of the proposed model, the standalone polynomial prior, and LSTM baselines.}
  \label{fig:section43_main_representative_trajectory}
\end{figure}

Figs.~\ref{fig:section43_main_representative_velocity} and~\ref{fig:section43_main_representative_trajectory} show representative predictions for two turning maneuvers and two zigzag maneuvers. Overall, the proposed model gives the most stable and balanced predictions in both velocity and trajectory. This can be seen across different maneuvering conditions, rather than from a single case. The polynomial model captures the general maneuvering trend, but it is less able to represent nonlinear response details. Its errors also accumulate during standalone rollout. For example, it gives less accurate surge predictions in the two zigzag cases and a less accurate sway prediction in the 15$^\circ$/15$^\circ$ zigzag case. LSTM-DC can reproduce some local velocity trends, but its trajectory predictions are less stable and show a clear gap from the proposed model. LSTM-Raw gives larger drift and lower accuracy. This shows that directly learning current-affected measured velocities is a simple but weak recursive target, with limited generalization. These comparisons suggest that the proposed differentiable hybrid framework is more effective than using either a purely physics-based model or a purely data-driven model alone.

\begin{figure}[htbp]
  \centering
  \includegraphics[width=0.96\linewidth]{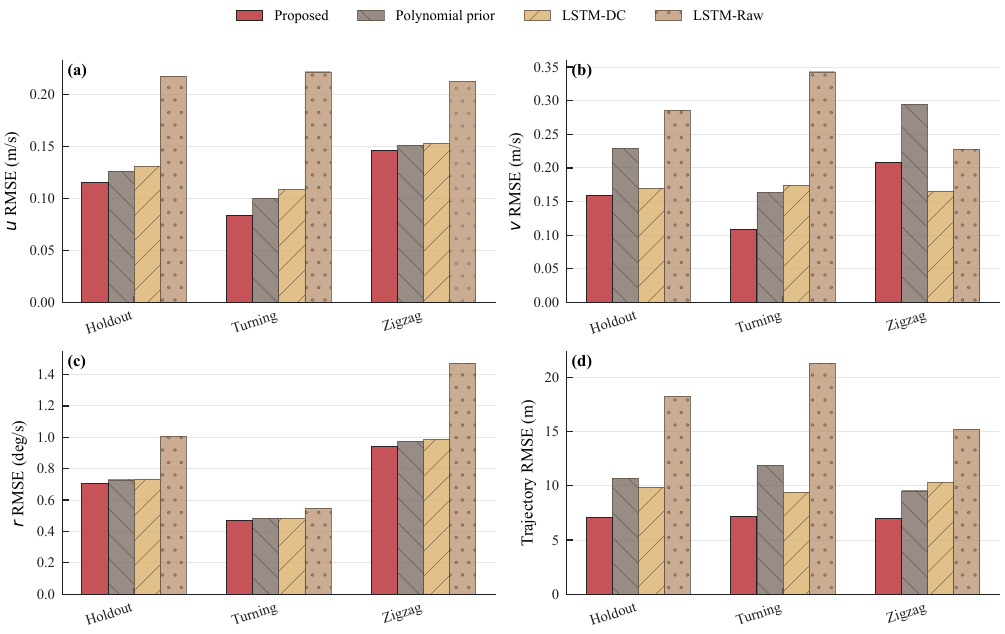}
  \caption{Channel-wise velocity RMSE and trajectory RMSE of the proposed model, the standalone polynomial prior, LSTM-DC, and LSTM-Raw on the holdout validation trials.}
  \label{fig:section43_main_summary}
\end{figure}

Fig.~\ref{fig:section43_main_summary} and Table~\ref{tab:comparison_summary} give the quantitative results, which are consistent with the visual comparison. The proposed model achieves the lowest errors in surge, sway, yaw, and trajectory prediction. This shows that the improvement is not limited to one response variable or one maneuver type. Instead, the model gives more balanced performance across all evaluated cases. This point is important because the main goal is not only to reduce the average error, but also to obtain stable and balanced generalization in recursive prediction. Such behavior is necessary for a model to be useful for reliable maneuvering forecast.

\begin{table}[htbp]
\centering
\caption{Recursive rollout results of the main comparison averaged over the eight holdout maneuvering trials}
\label{tab:comparison_summary}
{\scriptsize
\renewcommand{\arraystretch}{1.15}
\setlength{\tabcolsep}{2pt}
\begin{tabularx}{\textwidth}{>{\raggedright\arraybackslash}p{0.24\textwidth} *{4}{>{\centering\arraybackslash}X}}
\toprule
Model & \makecell{$u$ RMSE\\(m/s)} & \makecell{$v$ RMSE\\(m/s)} & \makecell{$r$ RMSE\\(deg/s)} & \makecell{Trajectory\\RMSE (m)} \\
\midrule
Proposed & 0.115 & 0.159 & 0.704 & 7.04 \\
Polynomial prior & 0.126 & 0.229 & 0.728 & 10.69 \\
LSTM-DC & 0.131 & 0.170 & 0.733 & 9.85 \\
LSTM-Raw & 0.217 & 0.285 & 1.007 & 18.21 \\
\bottomrule
\end{tabularx}
}
\end{table}

\begin{figure}[htbp]
  \centering
  \includegraphics[width=0.96\linewidth]{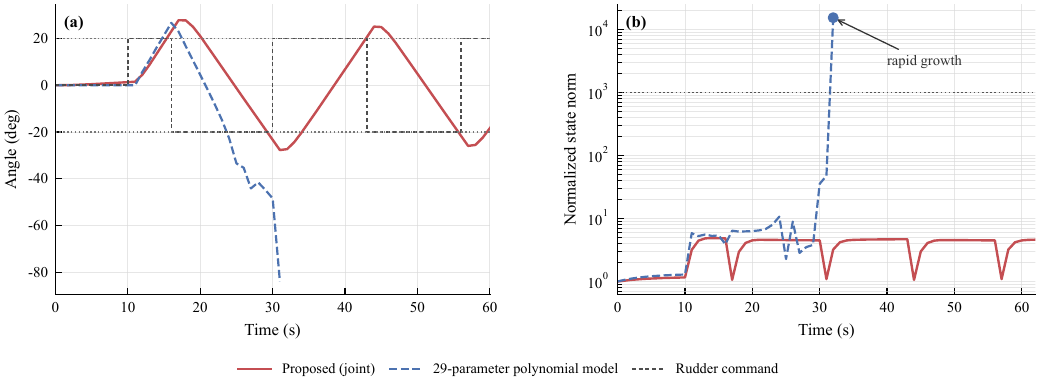}
  \caption{Standard 20$^\circ$/20$^\circ$ zigzag free-running prediction: (a) heading response and rudder command; (b) dimensionless normalized state norm for the proposed model and the 29-parameter polynomial maneuvering model.}
  \label{fig:section43_zigzag20_stability_comparison}
\end{figure}

As an additional diagnostic, Fig.~\ref{fig:section43_zigzag20_stability_comparison} compares the proposed model with a 29-parameter polynomial AUV maneuvering model following the structure used in \citep{hegrenaes2007comparison}. The 29-parameter model is fitted to the same training data and then evaluated in a standard 20$^\circ$/20$^\circ$ zigzag free-running simulation. This case is not included in the holdout average because it is not a sea-trial replay test. Under the tested setting, the 29-parameter polynomial model shows rapid growth during recursive free simulation, whereas the proposed predictor remains bounded within the plotted horizon. A possible explanation is multicollinearity among the polynomial basis functions, which may not be sufficiently and independently excited by the available maneuvers. Under these conditions, measurement noise can induce large, mutually compensating coefficient changes, yielding a reasonable fit to the calibration data without ensuring physically reliable coefficients \citep[p.~148]{ittc2008manoeuvring}. During recursive simulation, these compensations may no longer hold as predicted states depart from the calibration range, and feeding predictions back into the nonlinear polynomial model can further amplify the resulting deviations. 

This result indicates that the polynomial model structure must balance expressive ability and parameter identifiability. In the original reference model \citep{hegrenaes2007comparison}, lower and upper bounds were imposed on the parameters to be identified; in practical engineering applications, however, setting such bounds is not straightforward. In contrast, the proposed method uses a compact third-order polynomial prior and assigns the remaining discrepancy to a data-adaptive basis module, which improves the stability of the overall executable predictor in this diagnostic case.

To evaluate the stability of the model’s recursive predictions, multiple starting points were selected from the eight validation trials for two tests: 61 regularly spaced starting points and 16 starting points characterized by large absolute sway velocities or yaw rates. For each starting point, the model is provided with the required initial states and control inputs, and then recursively predicts surge velocity, sway velocity, yaw rate, and trajectory over 30 steps (30 s) using the recorded subsequent control sequence.

Table~\mbox{\ref{tab:measured_multistart_30step}} compares the resulting surge, sway, yaw-rate and trajectory errors. The proposed model gives the lowest mean RMSE in all four metrics for both start sets, with trajectory RMSEs of 1.932 m and 1.974 m, respectively. These results support consistent recursive prediction across the evaluated measured initial conditions, including larger sway velocities and yaw rates.

\begin{table}[htbp]
\centering
\caption{30-step recursive prediction errors from multiple measured initial states.}
\label{tab:measured_multistart_30step}
{\small
\renewcommand{\arraystretch}{1.16}
\setlength{\tabcolsep}{5pt}
\begin{tabularx}{\linewidth}{@{}>{\raggedright\arraybackslash}Xrrrr@{}}
\toprule
Model & \multicolumn{1}{c}{\makecell{\mbox{$u$} RMSE\\(m/s)}} & \multicolumn{1}{c}{\makecell{\mbox{$v$} RMSE\\(m/s)}} & \multicolumn{1}{c}{\makecell{\mbox{$r$} RMSE\\(deg/s)}} & \multicolumn{1}{c}{\makecell{Trajectory\\RMSE (m)}} \\
\midrule
\multicolumn{5}{@{}l}{\textit{A. Regularly spaced measured starts (61 starts)}} \\
\addlinespace[2pt]
Polynomial prior & 0.1060 & 0.1466 & 0.673 & 2.626 \\
Frozen-prior hybrid & 0.0997 & 0.1278 & 0.679 & 2.177 \\
Proposed & \textbf{0.0937} & \textbf{0.1264} & \textbf{0.648} & \textbf{1.932} \\
\midrule
\multicolumn{5}{@{}l}{\textit{B. Measured high-sway/high-yaw-rate starts (16 starts)}} \\
\addlinespace[2pt]
Polynomial prior & 0.1132 & 0.1683 & 0.703 & 3.192 \\
Frozen-prior hybrid & 0.1059 & 0.1240 & 0.712 & 2.541 \\
Proposed & \textbf{0.0979} & \textbf{0.1187} & \textbf{0.673} & \textbf{1.974} \\
\bottomrule
\end{tabularx}
\par\vspace{4pt}
\begin{minipage}{\linewidth}
\footnotesize
\end{minipage}
}
\end{table}

\subsubsection{\texorpdfstring{Ablation and sensitivity analysis}{Ablation and sensitivity analysis}}
\label{sec:calibration_ablation}
This subsection evaluates the contributions of the proposed joint calibration mechanism and current compensation through ablation studies, and examines the influence of calibration hyperparameters on recursive prediction. Two ablation models are used for comparison. Frozen-prior hybrid model uses the same current-compensated training data and data-adaptive basis structure as the proposed model, but keeps the polynomial coefficients fixed after the initial fit. Hybrid-Raw model directly used the measured velocity $\boldsymbol{\nu}_m$ as the training target, without current compensation. The two ablation models share the same composite structure as the proposed model, but they test the effect of making the polynomial prior trainable and the effect of training on current-affected data, respectively. 

\begin{figure}[htbp]
  \centering
  \includegraphics[width=0.96\linewidth]{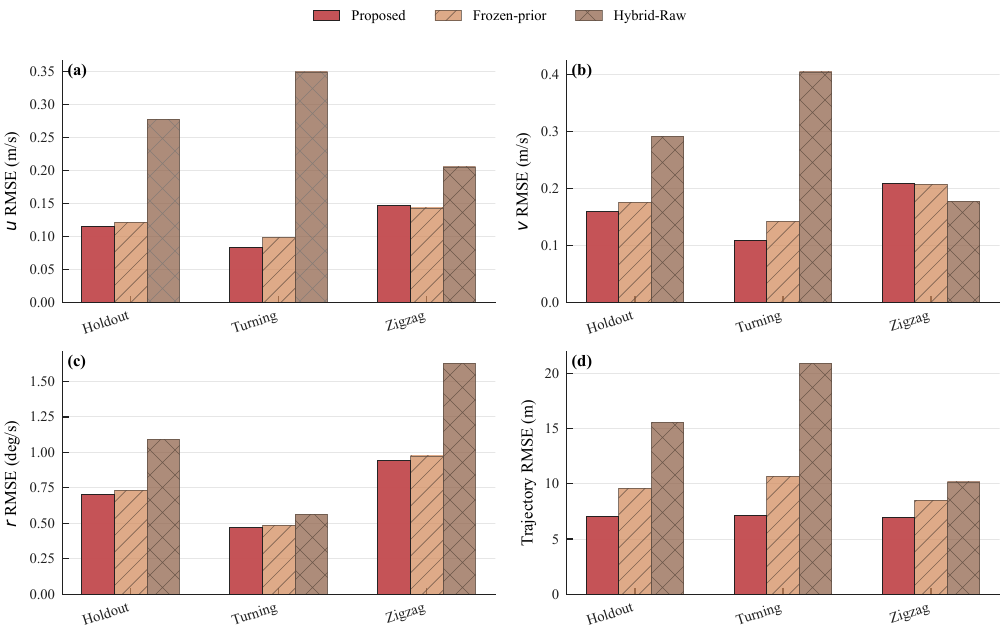}
  \caption{Ablation comparison of velocity RMSE and trajectory RMSE for trainable hydrodynamic-prior coefficients and current-compensated learning.}
  \label{fig:section43_ablation_summary}
\end{figure}

Fig.~\ref{fig:section43_ablation_summary} shows the ablation comparison results for the velocity components and trajectory prediction. Compared with the frozen-prior hybrid model, the proposed model achieves lower average RMSE values for $u$, $v$, $r$, and trajectory prediction, and yields lower trajectory RMSE in five of the eight holdout cases. These results suggest that bounded adjustment of the prior parameters enables the mechanistic model to absorb part of the systematic correction, rather than forcing the LSTM branch to compensate for both prior-model bias and residual dynamics. Although the frozen-prior model remains competitive, indicating that the LSTM branch can partially compensate for these errors, the proposed bounded co-calibration strategy provides a more interpretable and physically consistent error decomposition: low-frequency systematic corrections are assigned to the structured prior, while the LSTM branch focuses on the remaining nonlinear residual dynamics.

\begin{figure}[htbp]
  \centering
  \includegraphics[width=0.98\linewidth]{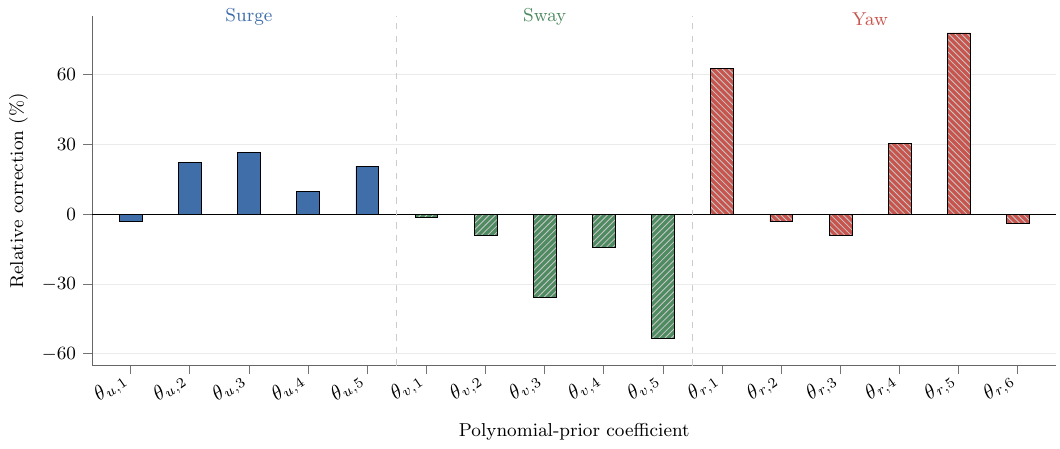}
  \caption{Relative correction of the effective hydrodynamic-prior coefficients in the selected joint model with $\rho=0.9$. The x-axis follows the coefficient notation in Eq.~\eqref{eq:polynomial_prior_basis}.}
  \label{fig:section43_coefficient_calibration_rho09}
\end{figure}

Fig.~\ref{fig:section43_coefficient_calibration_rho09} further illustrates the coefficient changes before and after bounded parameter perturbation. The corrections are selective rather than uniformly distributed, with a mean absolute correction of 24.0\% and a maximum correction of 77.9\%, occurring in a yaw-related coefficient. The result shows that the polynomial module remains active in the final predictor as a trainable structured component.

\begin{figure}[htbp]
  \centering
  \includegraphics[width=0.92\linewidth]{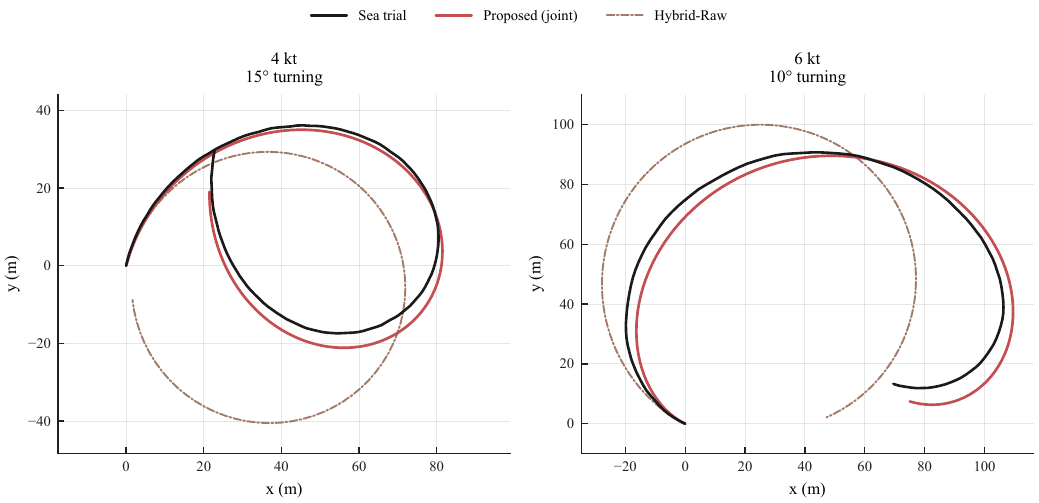}
  \caption{Recursive trajectory predictions in two turning maneuvers for the proposed current-compensated hybrid model and Hybrid-Raw.}
  \label{fig:section43_current_comp_turning_trajectory}
\end{figure}

Regarding current compensation within the differentiable composite model, Fig.~\ref{fig:section43_ablation_summary} and Fig.~\ref{fig:section43_current_comp_turning_trajectory} provide consistent evidence. Although the two models adopt the same model structure, Hybrid-Raw shows larger trajectory drift, indicating that the current projection is absorbed into the learned transition dynamics. These results support using body-relative, current-compensated velocity as the learning target, with the estimated current reintroduced only during rollout reconstruction.

\begin{figure}[htbp]
  \centering
  \includegraphics[width=0.98\linewidth]{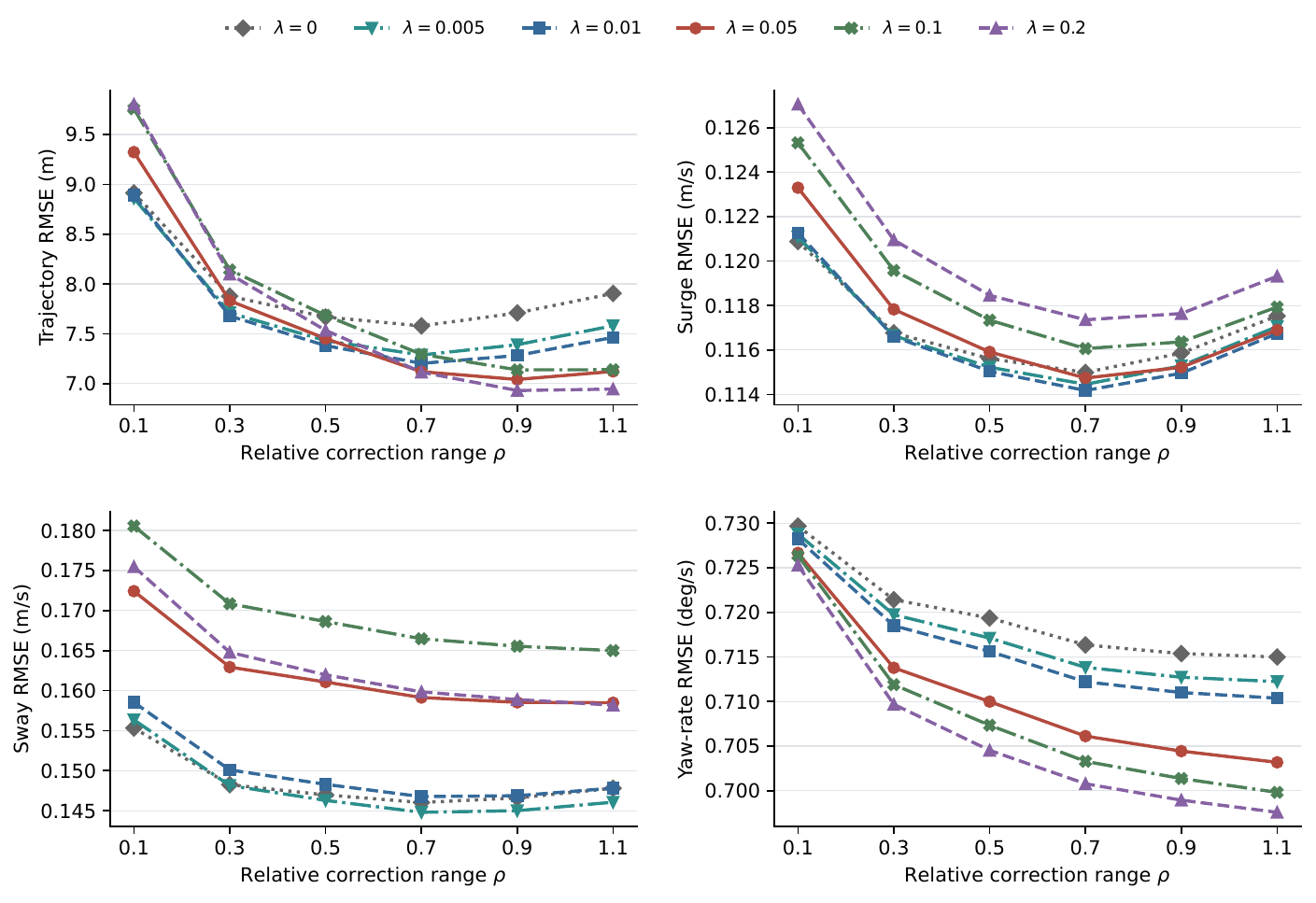}
  \caption{Recursive prediction sensitivity to the coefficient correction range \mbox{$\rho$} and adaptive-correction penalty \mbox{$\lambda$}.}
  \label{fig:section43_calibration_sensitivity}
\end{figure}

\FloatBarrier
Finally, a hyperparameter sensitivity test was conducted by varying the coefficient correction range \mbox{$\rho$} and the adaptive-correction penalty \mbox{$\lambda$}, with the other settings fixed. Fig.~\mbox{\ref{fig:section43_calibration_sensitivity}} shows the resulting velocity and trajectory prediction errors. Increasing \mbox{$\rho$} from small values generally improves prediction, with relatively modest changes over \mbox{$\rho=0.7$--$1.1$}. Larger \mbox{$\lambda$} generally reduces yaw-rate errors, but does not consistently improve the other prediction metrics. These results suggest that useful settings can be selected within a relatively broad parameter region in the present case, without precise tuning to a unique optimum.

\FloatBarrier

\subsection{Discussion}
The proposed framework provides an approximation-theoretic interpretation of hybrid maneuvering modeling through the composition of complementary basis components. A polynomial hydrodynamic model uses prescribed basis functions, such as velocity, rudder, propeller, and coupling terms, and then fits their coefficients from data. A data-driven recurrent model learns data-adaptive basis features from the maneuvering history. In this sense, a hybrid model is a composite approximation built from prescribed polynomial bases and learned data-adaptive bases.

The role of the unified differentiable architecture is to optimize these basis components under one prediction objective. The polynomial coefficients and the LSTM parameters are not calibrated in two isolated stages. They are placed in the same computational graph and updated by gradient descent. This provides additional flexibility in allocating the predicted response between the structured and adaptive components. This view also suggests that the framework is not limited to the particular basis used in this paper. Other differentiable basis components, coupling architectures, and parameter-update schemes can be incorporated within the same description, provided that the resulting predictor supports gradient-based calibration and executable recursive rollout. In the limiting case of a zero polynomial update range, the formulation recovers the corresponding frozen-prior residual-learning model within the same architecture.

Joint calibration also introduces a trade-off. The correction range, relative learning rates, and residual-regularization strength require problem-dependent selection. Appropriate settings depend on the approximation capacity and initial quality of the polynomial component, as well as the available data. Joint prediction optimization also does not automatically guarantee a unique separation of the polynomial and adaptive contributions. Accordingly, the calibrated coefficients should be interpreted as effective prediction coefficients rather than independently identified hydrodynamic derivatives.

Current treatment is another necessary part of the modeling problem. The polynomial maneuvering basis describes body-relative hydrodynamic response, whereas field measurements may contain current-induced kinematics. If the current component is not separated, the learned transition mixes vehicle dynamics with an environment-dependent projection. The current-compensated training target used here reduces this ambiguity and helps obtain an inherent maneuvering model.

The current validation has several limitations. First, it is conducted on a single type of AUV, and the generality of the proposed method should therefore be further assessed on additional platforms. From a methodological perspective, however, the formulation is extendable, and the full-scale experimental results reported in this study provide supporting evidence for its practical applicability. Second, the data were obtained from onboard measurements, including an inertial measurement unit and a Doppler velocity log. Although these measurements provide the basis for model calibration and validation, finite sensor accuracy and residual preprocessing biases are unavoidable. Within this scope, the model remains suitable for dead-reckoning support, maneuvering simulation, and digital-twin construction, where stable recursive prediction is more relevant than assigning a unique physical interpretation to each fitted coefficient. Further uncertainty analysis would be valuable for quantifying the effects of measurement errors, current compensation, and cross-platform variability.

\section{Conclusions}
This paper presented a differentiable composite approximation framework for AUV maneuvering prediction from full-scale sea-trial data. A reduced polynomial hydrodynamic prior and a data-adaptive basis module were embedded in one differentiable one-step transition map. The effective polynomial coefficients and neural parameters were optimized jointly, while ocean-current effects were removed during training and reintroduced during rollout.

Validation on eight holdout turning and zigzag maneuvers from full-scale AUV trials in real sea conditions shows that the proposed model improves average recursive prediction compared with the standalone polynomial prior and neural-only LSTM baselines. The results indicate that the polynomial prior and the data-adaptive basis correction play complementary roles in long-horizon rollout. The ablation results further show that bounded prior calibration and current-compensated learning both contribute to the final rollout performance. Future work should examine richer excitation, repeated-seed robustness, multi-step training objectives, and closed-loop validation.

\section*{CRediT authorship contribution statement}
\textbf{Aobo Wang:} Writing -- original draft, Methodology, Data curation.
\textbf{Aifei Xia:} Validation, Methodology, Formal analysis.
\textbf{Zihao Wang:} Writing -- original draft, Methodology, Formal analysis, Conceptualization.
\textbf{Lizhu Hao:} Writing -- review \& editing, Project administration, Funding acquisition.
\textbf{Shaorong Xie:} Formal analysis, Validation, Writing -- review \& editing.

\section*{Declaration of competing interest}
The authors declare that they have no known competing financial interests or personal relationships that could have appeared to influence the work reported in this paper.

\section*{Data availability}
The data that has been used is confidential.

\bibliographystyle{cas-model2-names}
\bibliography{cas-refs}

\end{document}